\documentclass[letterpaper]{article} % DO NOT CHANGE THIS

\ifdefined\aaaianonymous
    \usepackage[submission]{aaai2026}  % Anonymous submission version
\else
    \usepackage{aaai2026}              % Camera-ready version
\fi

\usepackage{times}  % DO NOT CHANGE THIS
\usepackage{helvet}  % DO NOT CHANGE THIS
\usepackage{courier}  % DO NOT CHANGE THIS
\usepackage[hyphens]{url}  % DO NOT CHANGE THIS
\usepackage{graphicx} % DO NOT CHANGE THIS
\usepackage{natbib}  % DO NOT CHANGE THIS AND DO NOT ADD ANY OPTIONS TO IT
\usepackage{caption} % DO NOT CHANGE THIS AND DO NOT ADD ANY OPTIONS TO IT
\usepackage{algorithm}
\usepackage{pdfpages} % Added to support \includepdf
\usepackage{algorithmic}
\usepackage{amsmath}
\usepackage{amssymb}
\usepackage{booktabs}
\usepackage{multirow}

\usepackage{newfloat}
\usepackage{listings}
\DeclareCaptionStyle{ruled}{labelfont=normalfont,labelsep=colon,strut=off} % DO NOT CHANGE THIS
\floatstyle{ruled}
\newfloat{listing}{tb}{lst}{}
\floatname{listing}{Listing}

\ifdefined\aaaianonymous
    \title{PerTouch: VLM-Driven Agent for Personalized and Semantic Image Retouching}
\else
    \title{PerTouch: VLM-Driven Agent for Personalized and Semantic Image Retouching}
\fi

\author{
    Zewei Chang\textsuperscript{\rm 1},
    Zheng-Peng Duan\textsuperscript{\rm 1},
    Jianxing Zhang\textsuperscript{\rm 2},
    Chun-Le Guo\textsuperscript{\rm 1,4},
    Siyu Liu\textsuperscript{\rm 1}, \\
    Hyungju Chun\textsuperscript{\rm 3},
    Hyunhee Park\textsuperscript{\rm 3},
    Zikun Liu\textsuperscript{\rm 2},
    Chongyi Li\textsuperscript{\rm 1,4}\thanks{Corresponding author.}
}
\affiliations{
    \textsuperscript{\rm 1}VCIP, CS, Nankai University\\
    \textsuperscript{\rm 2}Samsung R\&D Institute China - Beijing (SRC-B)\\
    \textsuperscript{\rm 3}The Department of Camera Innovation Group, Samsung Electronics\\
    \textsuperscript{\rm 4}NKIARI, Shenzhen Futian\\
    changzewei@mail.nankai.edu.cn, adamduan0211@gmail.com, \\
    \{guochunle, liusiyu29, lichongyi\}@nankai.edu.cn, %\\
    \{jx2018.zhang, hyungju.chun, inextg.park, zikun.liu\}@samsung.com
}

\begin{document}

\maketitle
\begin{abstract}
Image retouching aims to enhance visual quality while aligning with users' personalized aesthetic preferences. 
To address the challenge of balancing controllability and subjectivity, we propose a unified diffusion-based image retouching framework called \textbf{PerTouch}. 
Our method supports semantic-level image retouching while maintaining global aesthetics. 
%
% Focusing on parameter control, PerTouch constructs an explicit parameter-to-image mapping process and introduces semantic replacement and parameter perturbation mechanisms in the training process to enable semantic region-aware and fine-grained image retouching. 
Using parameter maps containing attribute values in specific semantic regions as input, PerTouch constructs an explicit parameter-to-image mapping for fine-grained image retouching. 
To improve semantic boundary perception, we introduce semantic replacement and parameter perturbation mechanisms during training. 
%
% To bridge natural language instructions and visual control, we further design an Agent that supports instruction following, a feedback-driven rethinking mechanism for interpreting vague expressions, and a scene-aware memory mechanism to model long-term user preferences. 
To connect natural language instructions with visual control, we develop a VLM-driven agent to handle both strong and weak user instructions. 
Equipped with mechanisms of feedback-driven rethinking and scene-aware memory, PerTouch better aligns with user intent and captures long-term preferences. 
%
% PerTouch can understand subjective instructions, adapt to diverse image semantics, and produce high-quality results consistent with users' long-term editing styles. 
%
Extensive experiments demonstrate each component’s effectiveness and the superior performance of PerTouch in personalized image retouching. 
\end{abstract}

% Links section - only shown in camera-ready version
% \ifdefined\aaaianonymous
% NOTE: For anonymous submissions, do not include links that could reveal your identity
\begin{links}
    \link{Code}{https://github.com/Auroral703/PerTouch}
    % \link{Datasets}{https://aaai.org/example/datasets}
    % \link{Extended version}{https://aaai.org/example/extended-version}
\end{links}

\section{Introduction}

\begin{figure*}[!t]
    \centering
    \vspace{-10pt}
    \includegraphics[width=0.85\textwidth]{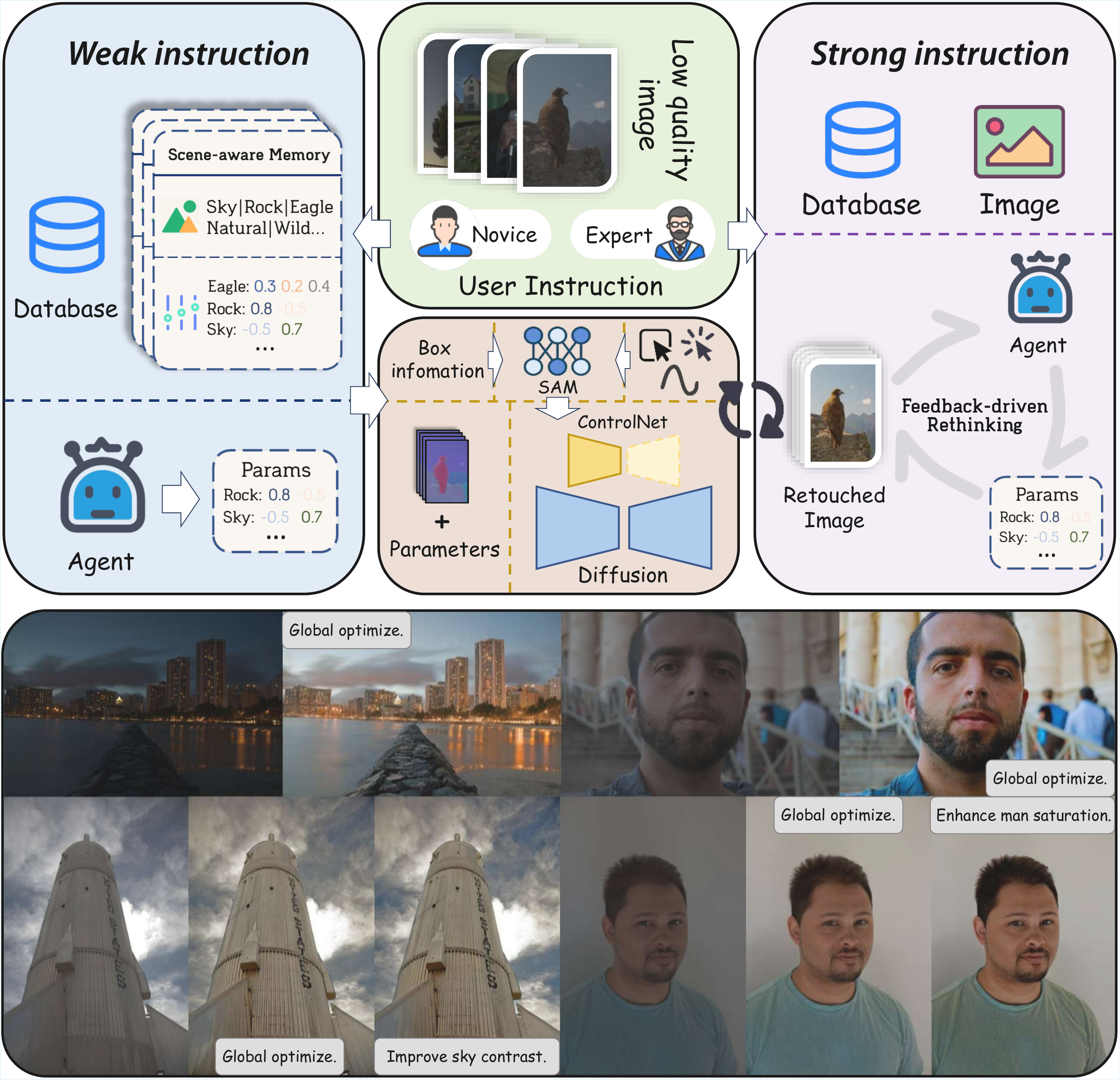}
    \setlength{\abovecaptionskip}{8pt}
    \caption{Overview of our \textbf{PerTouch} pipeline. Our method supports region-level personalized retouching with long-term user memory. Given images and natural language instruction, PerTouch determines the strength of the instruction, and then leverages the scene-aware memory to adaptively perform corresponding retouching operations based on the user's historical preferences. The final result is retouched by a fine-tuned diffusion model, ensuring globally pleasing and finely controlled region-level edits. The examples at the bottom demonstrate the system's ability to perform both global retouching and fine-grained regional adjustment across various instruction types.}
    \label{fig:Overview}
    \vspace{-10pt}
\end{figure*}

With the increasing accessibility of photography devices and the lowering threshold for photo-taking, capturing images has become an essential medium for personal expression. 
However, due to the lack of professional photography knowledge and the uncontrollable shooting environment, raw photos often fail to achieve satisfactory visual quality. 
To bridge this gap, image post-processing has become a crucial technique for enhancing photo quality and improving visual expressiveness. 
While professional software such as Adobe Lightroom \cite{AdobeLightroom} and Photoshop \cite{AdobePhotoshop} provides powerful tools for image retouching, these systems typically require expert knowledge and involve complex workflows, making them less accessible to ordinary users, especially for batch processing or personalized style editing at scale.

Therefore, a number of deep learning-based image retouching methods have been proposed.
Yet, existing approaches still face several limitations. 
%
% First, different users have diverse aesthetic preferences, and their intentions are often highly subjective. 
% %
% Non-expert users usually struggle to describe their editing needs precisely, often providing vague or abstract instructions. 
% %
% These inherently human expressions actually reflect the need for fine-grained, region-aware retouching, which is often overlooked by existing methods. 
%
The limitations of current approaches mainly fall into three categories. (1) \textbf{Lack of subjectivity modeling}: 
Most methods adopt deterministic architectures that generate a single fixed result for a given input, failing to account for the diversity and subjectivity of user preferences.
(2) \textbf{Lack of region-level control}: While some works introduce controllable parameters and reference images to control the image retouching style \cite{duan2025diffretouch,oywq2023rsfnet}, they often fail to support flexible local editing. 
Attempts that incorporate external segmentation maps are sensitive to segmentation quality and tend to produce visually unnatural results.
(3) \textbf{Lack of user interaction modeling and personalization}: Existing methods typically cannot interpret vague user instructions and ignore the need to memorize long-term editing preferences, resulting in poor adaptability and high user burden, particularly in batch or repeated editing scenarios.

To address these challenges, we propose \textbf{PerTouch}, a unified framework for fine-grained and personalized image retouching. 
Our method leverages the powerful diffusion prior to learn a diverse and high-quality retouching distribution, enabling the generation of globally aesthetic yet regionally consistent images conditioned on user intent. 
To enable semantic-aware regional editing, we propose a novel data preprocessing strategy that incorporates semantic replacement and parameter perturbation during training, helping the model better perceive semantic boundaries and mitigate overfitting to segmentation information existing in inputs. 
Furthermore, we design an agent driven by a vision language model (VLM) to lower the barrier for user interaction. 
The agent supports both strong and weak natural language prompts and can interpret vague instructions by inferring parameters in context. 
In addition, we introduce a scene memory mechanism to record the user's editing preferences under different semantic scenarios, enabling personalized and context-aware retouching over long-term usage.

In summary, our contributions are as follows:
\begin{itemize}
  \item We propose a semantic-aware region adjustment strategy based on diffusion priors, enabling globally aesthetic and locally consistent image retouching.
  \item We design a data preprocessing scheme combining semantic replacement and parameter perturbation to improve semantic boundary perception and parameter learning.
  \item We develop a VLM-driven agent with a scene memory mechanism to model long-term user preferences as well as enable personalized and context-aware retouching.
\end{itemize}

% Extensive experiments on standard benchmarks demonstrate that \textbf{PerTouch} can produce visually pleasing, semantically reasonable, and user-aligned retouching results.

\section{Related Work}

\subsection{Image Retouching in Deep Learning}

Recent advances in deep learning, alongside the availability of high-quality datasets \cite{5995332,jie2021PPR10K}, have driven significant progress in automated image retouching. 
Early approaches predominantly adopt Fully Convolutional Networks (FCNs) for end-to-end image-to-image translation \cite{Chen2018LearningTS,8578758,He2020ConditionalSM,9710400,Sun2021EnhanceIA}, while others incorporate photographic priors such as Retinex theory \cite{9577287,9102962,Wang_2019_CVPR}, 3D-LUTs \cite{Zeng2020LearningI3,Yang2022AdaIntLA,Wang2021RealtimeIE}, or curve and grid-based operations \cite{moran2020curl,Gharbi2017DeepBL,Moran_2020_CVPR,song2021starenhancer} to enhance interpretability and controllability. 
To address aesthetic diversity, style transfer methods \cite{kim2020pienet,10225702,song2021starenhancer} enable multi-style outputs but often rely on reference exemplars, which increases user burden. More recently, diffusion-based methods such as DiffRetouch \cite{duan2025diffretouch} have shown promise in modeling the complex distribution of expert-retouched styles via interpretable attribute control. 
However, most existing methods lack regional controllability or overly depend on external masks \cite{oywq2023rsfnet}, which may lead to unnatural artifacts. 
To this end, we propose PerTouch which enables semantic-aware regional retouching while preserving global aesthetic quality. 
Our method introduces explicit region-to-parameter mapping and supports fine-grained control and user interaction, addressing the limitations of both deterministic and reference-driven approaches.

\subsection{Agent in Low-level Vision}

With the rise of vision language models, researchers have begun leveraging their strong visual priors to perceive and invoke external tools, driving substantial progress in agent-based systems for low-level vision. Recent studies \cite{li2025hybridagents,chen2024restoreagent,zhu2024agenticir,Jiang2025MultiAgentIR} employ agents to tackle various degradation tasks toward achieving all-in-one restoration capabilities. Parallel to this, a line of work \cite{chen2025photoartagent,jarvisart2025,dutt2025monetgpt} explores using vision language models as agents to interact with image retouching toolchains like Lightroom, enabling automated photo retouching through exposure, contrast, and tone curve adjustments guided by language instructions. However, current retouching systems typically rely on fixed tool invocation pipelines and lack adaptability to individual user preferences. To address this, we propose a scene memory mechanism that stores users’ editing history, infers personalized preferences, and generates retouching results aligned with user intent, enabling truly personalized and preference-aware photo retouching.

\section{Methodology}

\subsection{Overview}
\label{sec:overview}
Given a low-quality input image $X$, the objective of image retouching is to generate a high-quality image that aligns with human aesthetic preferences while preserving original details.
%
% To address the previously underexplored challenges of semantic-aware, region-level adaptive retouching and personalized retouching based on user preferences, we propose PerTouch.
We propose PerTouch, a diffusion-based approach to address the underexplored challenges of semantic-aware, region-level adaptive retouching and personalized enhancement based on user preferences.
An overview of our framework is illustrated in Figure~\ref{fig:Overview}.
Similar to DiffRetouch \cite{duan2025diffretouch}, to assist users in retouching image styles that match their aesthetics, we provide four predefined image attributes (colorfulness, contrast, color temperature, and brightness) to facilitate intuitive user control. 
Our method is extensible: once a region-level score can be computed for a new attribute, our framework can incorporate it to enable controllability over that attribute. 
%
% To fully leverage the diffusion prior, we adopt Stable Diffusion \cite{Rombach2021HighResolutionIS} as the backbone and introduce ControlNet \cite{zhang2023adding} to inject region-level attribute information, thereby enabling control over the generation process. 
To fully leverage the diffusion prior, we adopt Stable Diffusion \cite{Rombach2021HighResolutionIS} as the backbone and introduce ControlNet \cite{zhang2023adding} to inject region-level attribute information, enabling control over the generation process.
Detailed model architecture is presented in Section \ref{sec:architecture}.

To facilitate user interaction with the system and preserve editing preferences across users, we introduce a VLM-based agent. 
We categorize user instructions into strong instructions and weak instructions, and invoke the retouching algorithm process accordingly. 
To incorporate users’ historical editing preferences, we store scene-aware memory for each editing session, which helps guide the agent’s decision-making. 
Furthermore, we design a feedback-driven rethinking mechanism to help the agent understand the relationship between parameter changes and image variations, enabling multi-stage decision-making that leads to results more aligned with user intent. 
The agent design and workflow is detailed in Section \ref{sec:agent}.

\subsection{Architecture}
\label{sec:architecture}
\begin{figure*}[t]
    \centering
    \vspace{-14pt}
    \hspace*{-0.02\textwidth}
    \includegraphics[width=0.92\textwidth]{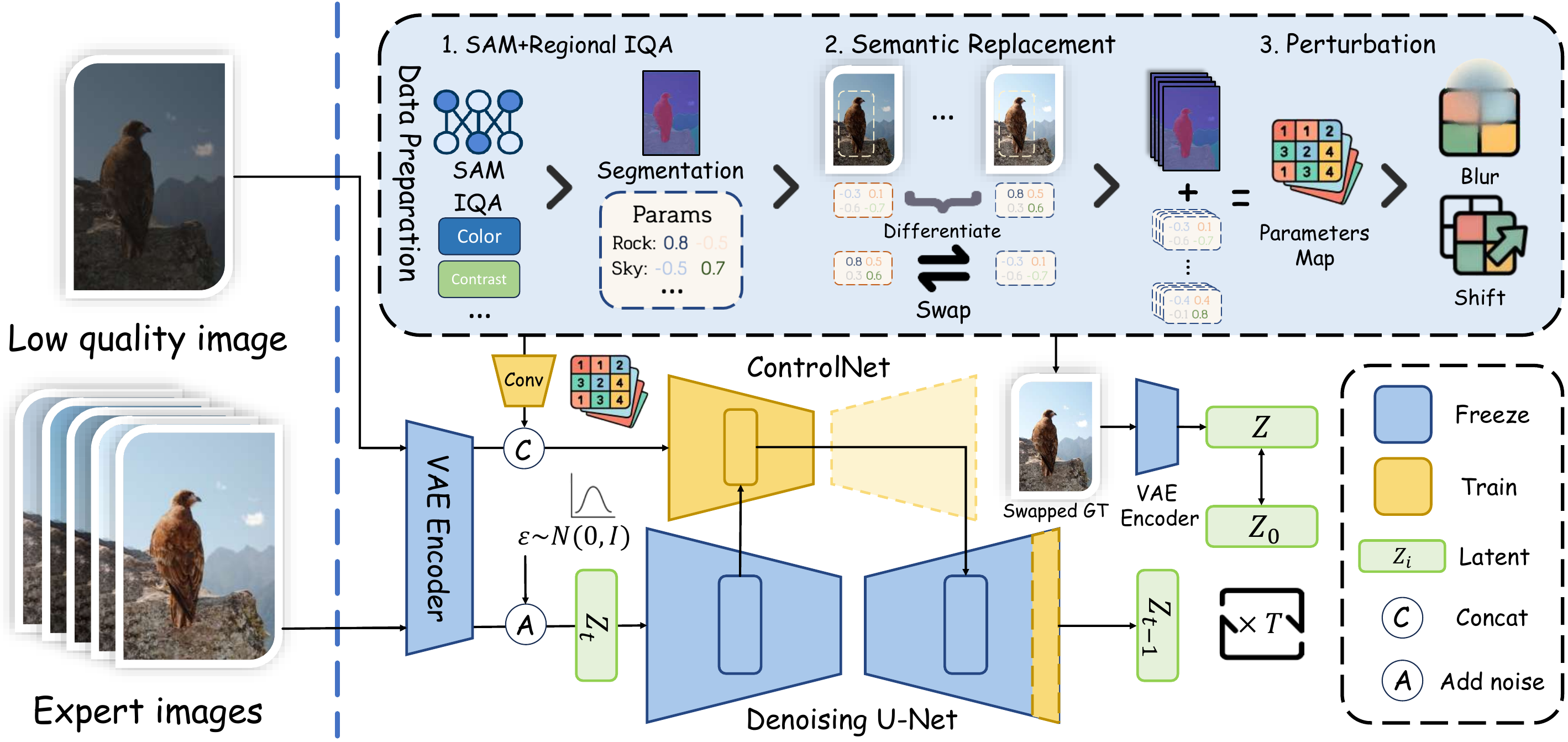}
    \setlength{\abovecaptionskip}{6pt}
    \caption{Dataset construction and training pipeline of PerTouch. To enable region-level controllable retouching, we construct training samples by generating parameter maps that transform low-quality input images into expert-retouched ground truth results. Specifically, we 1. extract semantic masks using SAM and estimate corresponding attribute parameters for each region; 2. introduce the Semantic Replacement Module to help the model perceive semantic regions by constructing diverse yet semantically consistent samples; and 3. apply the Perturbation Mechanism to prevent overfitting to segmentation boundaries and improve overall visual quality. The final parameter maps are injected into ControlNet alongside the original images, enabling the model to balance the global aesthetic consistency provided by diffusion priors and the regional guidance from parameter maps, thereby producing high-quality region-aware retouching outputs.}
    \label{fig:Train}
    \vspace{-10pt}
\end{figure*}

\subsubsection{Data Preparation}

Our primary dataset is the MIT-Adobe FiveK dataset, which consists of 5,000 RAW images, each accompanied by five expert-retouched reference versions (A/B/C/D/E). 
Additional dataset details are presented in the Supplementary Material. %Section \ref{sec:settings}. 
To enable the model to learn the relationship between regional attribute values and visual changes, we provide the model with paired data of parameter maps and corresponding images for supervised training. 

To obtain coarse semantic segmentation maps as auxiliary guidance, we leverage the panoptic segmentation capability of SAM. 
SAM automatically samples a set of evenly distributed points across the image and generates multiple segmented regions by treating each point as an independent prompt. 
A series of post-processing steps, including non-maximum suppression (NMS), is applied to obtain a high-confidence panoptic segmentation map. 
Once the segmentation map is obtained, we evaluate each semantic region using predefined regional scoring methods to assign attribute-specific scores. 
The segmentation and scoring information is then fused into a single parameter map by embedding the scores into the segmentation map and extending its channel dimensions to match the number of controllable attributes. 
This parameter map serves as guidance for attribute-aware image retouching.
The detailed data preparation pipeline is marked in blue in Figure~\ref{fig:Train}. 

We observe that directly injecting the parameter maps into the network leads to over-reliance on the information it encodes, which is contrary to our objective. 
Rather than enforcing strict adherence to the parameter maps, we aim to use them as a soft hint, allowing the diffusion prior to play a central role in generating aesthetically pleasing results. 
To mitigate the model's over-reliance on the parameter map, we introduce two mechanisms.

First, we find that the model struggles to perceive region boundaries based on the injected parameter map, often defaulting to global retouching. 
This is because the input map contains only coarse semantic scores, which are not spatially continuous across regions, while real images often exhibit spatially coherent color transitions. 
This discrepancy makes it difficult for the model to correlate the parameter maps with the image structure. 
To address this, we propose a semantic replacement module. During training, a small subset of samples is selected, and a region is randomly chosen based on semantic area size as probability. 
The selected region is replaced with a region from another sample with the most divergent attributes in the parameter space. 
This artificial manipulation encourages the model to detect regional discrepancies and thus develop fine-grained retouching capabilities.

Second, although the semantic replacement module facilitates local retouching, the model tends to ignore global coherence, resulting in visually inconsistent and aesthetically unpleasing outputs. 
This suggests that the model is overly sensitive to the segmentation information contained in the parameter maps and not fully utilizing the strong generative capacity of the diffusion prior. 
To alleviate this, we introduce perturbations to the parameter maps along multiple dimensions, such as channel shifts and blurring, thereby enforcing the treatment of the segmentation as soft guidance rather than a rigid structure. 
This encourages the model to interpret and respond to semantic boundaries implicitly during the generation process. 
The effectiveness of both modules is further examined in Section \ref{sec:ablation}.

\subsubsection{Baseline}

Our baseline model builds upon Stable Diffusion, which extends denoising diffusion probabilistic models (DDPM) by operating in a learned latent space rather than directly in pixel space. 
%
% A powerful autoencoder, consisting of an encoder \(E\) and a decoder \(D\), is pre-trained to map an input image \(X\) into its latent representation \(Z = E(X)\), and reconstruct the image from latent space via \(D(Z)\). 
%
The denoising model \(\varepsilon_\theta(Z_t, t, m)\) is trained to reverse the noise process in the latent space, where \(Z_t\) denotes the noised latent at timestep \(t\), and \(m\) represents conditioning signals. 
%
% In our setting, we focus on leveraging the powerful generative priors of the diffusion model without relying on additional semantic guidance. To this end, we use empty textual cues as conditions.

To enable fine-grained control over regional image attributes, we integrate ControlNet into the Stable Diffusion framework. 
%
% ControlNet augments the original time-conditional U-Net backbone with a parallel control branch $\mathcal{F}_{\text{ctrl}}$, which receives additional structural conditions $C \in \mathbb{R}^{H \times W \times d}$ to control generation. 
%
% This branch is composed of zero-initialized convolutional residual blocks that extract controllable features $\Delta h_t = \mathcal{F}_{\text{ctrl}}(C, t)$ at each timestep $t$. 
%
% The extracted control features are injected into multiple layers of the U-Net by residual addition with the main backbone features $h_t$, producing the guided representation:
% \begin{equation}
%     h_t' = h_t + \Delta h_t.
% \end{equation}
% During training, the parameters of the main denoising model $\theta$ are frozen to preserve the powerful diffusion prior, while only the control branch parameters $\phi$ are optimized. 
%
To accommodate multi-attribute conditioning, we expand the region-level attribute scores into a multi-channel guidance map $C = \{C^1, C^2, \dots, C^K\}$ of the same spatial resolution as the segmentation map, where each channel $C^k$ encodes the spatial distribution of a specific image attribute (e.g., colorfulness, contrast, color temperature, brightness). 
Each pixel value in $C^k$ reflects the score of its corresponding region in that attribute.
% This design enables the model to retain the global high-quality image synthesis capabilities of Stable Diffusion while achieving region-aware, attribute-guided photo retouching. 
%
By adjusting the values of specific semantic regions within \( C \), the model outputs corresponding retouching styles for the associated attributes, while simultaneously maintaining global image aesthetics. 
The coefficient values are adjusted within the range \([-1, 1]\), where each value corresponds to a visual style learned within the distribution of high-quality images. 
%
% For instance, in the "brightness" channel, a higher value tends to produce brighter styles, while a lower value corresponds to darker styles. 
%
% In this manner, the model can flexibly sample diverse yet aesthetically consistent image results under user guidance, enabling personalized region-level photo retouching.
The specific parameter maps injection method and training details are given in the Supplementary Material.

\subsection{Agent}
\label{sec:agent}
\begin{figure}[!ht]
    \centering
    \vspace{-13pt}
    \includegraphics[width=0.95\linewidth]{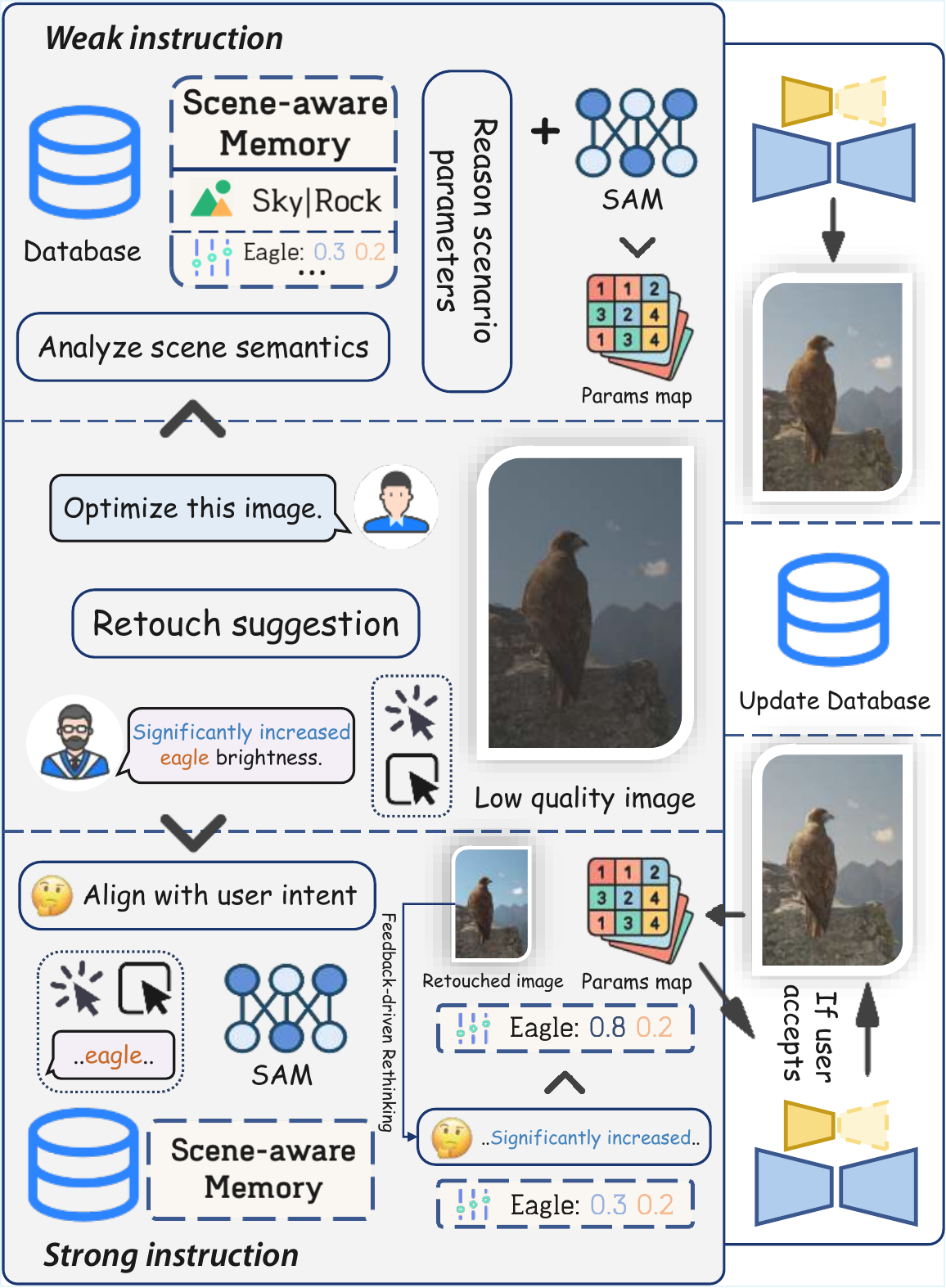}
    \setlength{\abovecaptionskip}{5pt}
    \caption{Agent workflow in PerTouch. Our unified agent framework adaptively parses user instructions of varying strength. For weak instructions (e.g., “Optimize this image.”), the agent leverages scene-aware memory to retrieve long-term user preferences and generates editable parameter maps based on historical behavior. For strong instructions (e.g., “Significantly increased
    eagle brightness.”), the agent further adopts a feedback-driven rethinking mechanism to iteratively refine vague or unsatisfactory outputs. This adaptive instruction-following pipeline allows PerTouch to support both global and region-level personalized retouching under natural language commands.}
    \label{fig:workflow}
    \vspace{-6pt}
\end{figure}

\subsubsection{Instruction Types and Agent Strategies}

To accommodate diverse user demands and editing intentions in personalized photo retouching, we design an interactive and preference-aware agent. 
Our agent supports two types of instruction parsing: weak instructions and strong instructions, aimed at simulating user needs ranging from casual, quick edits to professional, fine-grained retouching.

Weak instructions are designed for non-expert users who prefer minimal interaction. 
In this mode, the agent automatically constructs the multi-channel parameter maps $C = {C^1, C^2, \dots, C^K}$ using the midpoint of each image attribute as the default value. 
While some methods \cite{song2021starenhancer} allow users to provide reference images to indicate their preferred style, we adopt the midpoint-based initialization to reduce user learning costs and improve usability.
The parameter maps are then further tailored based on the user’s editing history and preferences.
This guidance map is then injected into the ControlNet to enable rapid and accurate region-aware retouching, producing visually appealing results that align with the user’s historical aesthetic tendencies, without requiring explicit manual input.

In contrast, strong instructions are intended for users with clearer editing goals or professional demands. 
In this mode, users can specify the target region, the attribute dimension(s) to be modified, and the desired retouching strength. 
Upon receiving instructions, the agent leverages the VLM’s object detection capability to identify the region of interest and invokes the SAM model to obtain a coarse segmentation mask. 
Based on this, a revised guidance map is constructed via a feedback-driven Rethinking mechanism, applying precise modifications to the designated region while retaining the adjustments derived from the weak instruction elsewhere. 
This design enables accurate local retouching while preserving global coherence and aesthetic consistency.

\subsubsection{Feedback-driven Rethinking}

In real-world usage scenarios, users typically lack a precise understanding of the parameter space. 
As a result, they often do not provide exact adjustment values, but instead use vague or subjective expressions such as “slightly increase”, “increase significantly”, or “reduce a bit”. 
This introduces a challenge for the model to interpret the intended degree of retouching and to determine whether the generated result aligns with the user’s instruction.

To address this, we propose a Feedback-driven Rethinking mechanism. 
At the initial stage, the model estimates a control value \( c_0 \) by conditioning on the user's instruction \( I \), the model's prior knowledge \( \mathcal{P} \), and the user's historical preferences \( \mathcal{H}_u \):
\begin{equation}
c_0 \sim p(c \mid I, \mathcal{P}, \mathcal{H}_u), \quad \hat{X}_0 = \mathcal{G}(X, c_0)
\end{equation}
Here, \( \hat{X}_0 \) denotes the first-round retouched image produced by the generative model \( \mathcal{G} \), and \( X \) is the original image. 
Importantly, the instruction \( I \) implicitly corresponds to an ideal control value \( c^* \), which would generate a preferred image \( X^* \). 
However, since \( X^* \) is not directly accessible, the system initiates an iterative rethinking process. 
At each step \( t \), the current output \( \hat{X}_t \) is sent to the agent alongside the original image \( X \) and the instruction \( I \), allowing the multi-modal model to assess whether the result satisfies the intended semantic adjustment. 
If not, the agent revises the control variable by incorporating feedback from the previous result:
\begin{equation}
c_t \sim p(c \mid I, \hat{X}_{t-1}, \mathcal{P}, \mathcal{H}_u), \quad \hat{X}_t = \mathcal{G}(X, c_t)
\end{equation}
This forms a closed loop of parameter refinement and visual retouching, where the system progressively adjusts the parameters to bring the output \( \hat{X}_t \) closer to the latent target \( X^* \) aligned with the user’s instructions. 
This mechanism not only enables the model to handle ambiguous user expressions, but also facilitates the construction of a learned mapping between language-level adjustment cues, control values, and perceptual visual outcomes. 
Ultimately, it improves the alignment between user intent and retouching results.

\begin{figure*}[!t]
    \centering
    \vspace{-10pt}
    \includegraphics[width=0.85\linewidth]{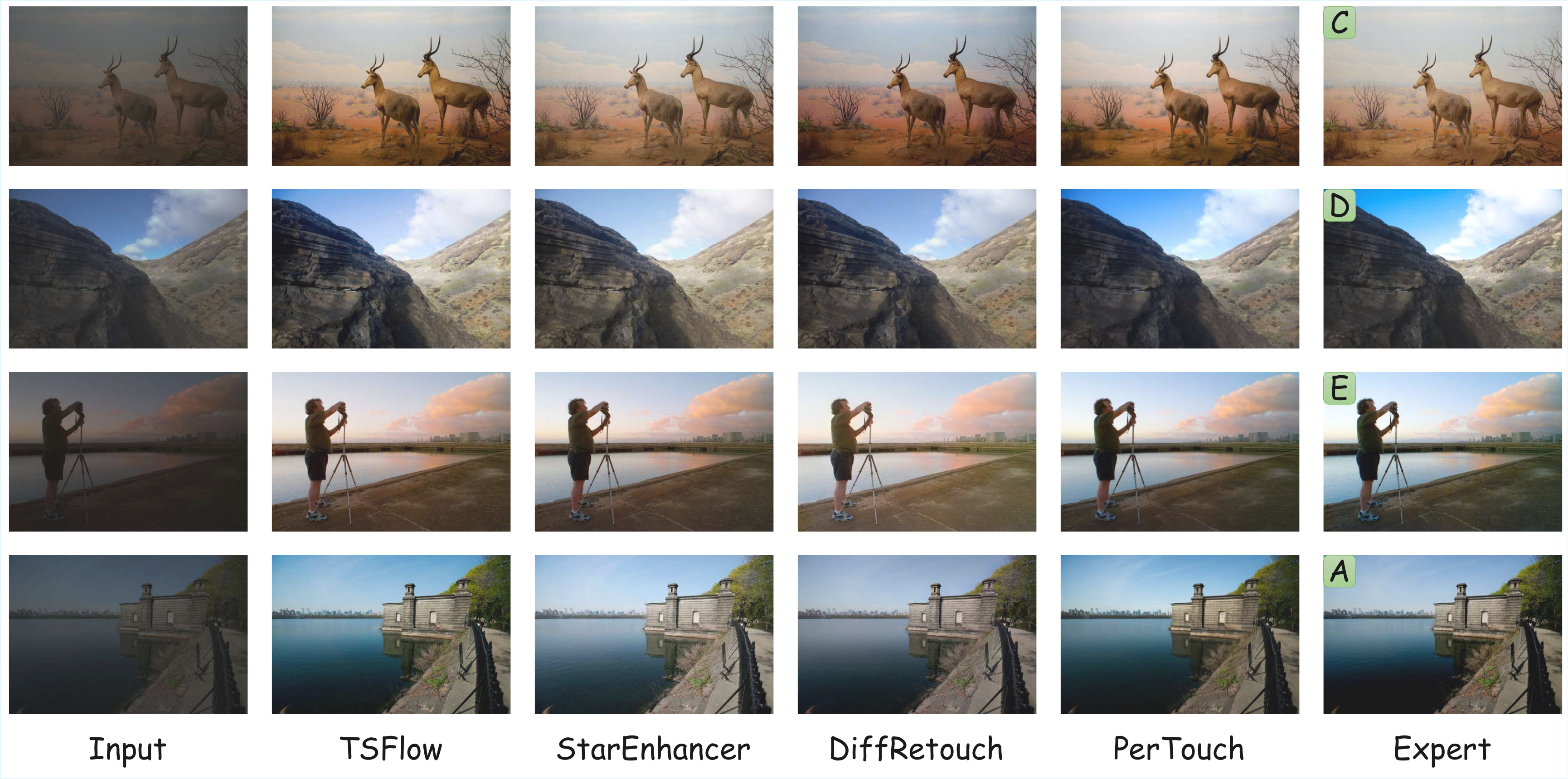}
    \setlength{\abovecaptionskip}{4pt}
    \caption{Qualitative comparison with other methods.}
    \label{fig:Comparisons}
\end{figure*}

\begin{table*}[!htb]
    \small
    \centering
    \vspace{-5pt}
    \renewcommand{\arraystretch}{1.3}
    \tabcolsep=0.18cm
    \begin{tabular}{@{}c|cc|cc|cc|cc|cc@{}}
\toprule
\multirow{2}{*}[-0.7ex]{\centering Method} & \multicolumn{2}{c|}{A} & \multicolumn{2}{c|}{B} & \multicolumn{2}{c|}{C} & \multicolumn{2}{c|}{D} & \multicolumn{2}{c}{E} \\
\cmidrule{2-11}
& PSNR~$\uparrow$ & LPIPS~$\downarrow$ & PSNR~$\uparrow$ & LPIPS~$\downarrow$ & PSNR~$\uparrow$ & LPIPS~$\downarrow$ & PSNR~$\uparrow$ & LPIPS~$\downarrow$ & PSNR~$\uparrow$ & LPIPS~$\downarrow$ \\
\midrule
PIENet       & 21.5184 & 0.1265 & 25.9065 & 0.0912 & 25.1927 & 0.0975 & 22.8989 & 0.1119 & 24.1171 & 0.1131 \\
TSFlow       & 20.6123 & 0.1037 & 25.2474 & 0.0716 & 25.6243 & \textbf{0.0630} & 22.3720 & 0.0894 & 23.5393 & 0.0822 \\
StarEnhancer & 20.7100 & 0.1057 & 25.7296 & 0.0738 & 25.5198 & 0.0645 & 23.3875 & \underline{0.0803} & 24.4558 & 0.0834 \\
Diffretouch  & \underline{24.5082} & \underline{0.0812} & \underline{26.1473} & \textbf{0.0672} & \underline{25.9148} & \underline{0.0684} & \underline{24.5087} & \textbf{0.0768} & \underline{24.7373} & \textbf{0.0776} \\
PerTouch     & \textbf{25.1430} & \textbf{0.0798} & \textbf{27.4733} & \underline{0.0687} & \textbf{26.7510} & 0.0844 & \textbf{25.9726} & 0.0823 & \textbf{25.6602} & \underline{0.0792} \\
\bottomrule
\end{tabular}
\setlength{\abovecaptionskip}{8pt}
\caption{Quantitative comparisons on the MIT-Adobe FiveK dataset. Evaluations are conducted on five expert retouching versions (A/B/C/D/E) in the test set, with each model provided the appropriate condition for generating expert-style outputs. Best results are shown in \textbf{bold}, and second-best are \underline{underlined}.
}
\label{tab:quantitative_comparison}
\vspace{-10pt}
\end{table*}

\subsubsection{Scene-aware Memory}

To further enhance the model’s capacity for capturing long-term user preferences, we introduce a mechanism called Scene-aware Memory. 
After the image retouching operation, the agent extracts the scene semantics of the image $\mathcal{F}(\cdot)$, yielding $f_t = \mathcal{F}(I_t)$, and stores them alongside the final confirmed editing parameter $\hat{c}_t$ in the personalized memory bank $\mathcal{M}$. 
As the number of user interactions increases, the memory gradually encodes a distribution of user preferences conditioned on different scenes.

When the user later edits a new image $I_q$, the agent first extracts its scene semantics $f_q = \mathcal{F}(I_q)$. 
Based on these semantics and the memory bank $\mathcal{M}$, the agent estimates a conditional preference distribution $p(c \mid f_q; \mathcal{M})$, and samples a parameter vector to guide the editing process:
\begin{equation}
\tilde{c}_q \sim p(c \mid f_q; \mathcal{M})
\end{equation}
The sampled parameter $\tilde{c}_q$ serves as the control signal for downstream image editing modules, enabling the model to generate outputs that better align with the user’s long-term aesthetic tendencies. 
This mechanism allows the system to maintain consistent personalization across diverse users and scene types, while significantly reducing the burden of manual parameter tuning and improving both interaction efficiency and visual coherence.

\section{Experiments}

\subsection{Settings}

Detailed dataset information and experimental settings are presented in the Supplementary Material.

\subsection{Comparisons}
\label{sec:comparison}
We compare our PerTouch with several existing image retouching methods, focusing on approaches that support diverse retouching styles. 
These include DiffRetouch, StarEnhancer, TSFlow and PIE-Net, which adopt a single model trained on retouched results from multiple experts. 
During inference, these models can generate style-specific outputs either by providing control parameters or extracting the style from reference images to emulate different expert preferences. 
To evaluate the multi-style retouching capability of PerTouch, we follow the same data preparation pipeline as in the training phase to generate expert-specific supervision on the test set. 
For each low-quality input image in the MIT-Adobe FiveK test set, we construct a set of expert guidance maps $C = \{C^1, C^2, \dots, C^K\}$ according to the procedure in Sec.~3.2, enabling the model to produce multiple expert-style outputs under different conditions. 
Since all compared methods support multi-style generation, we evaluate retouching results across all five experts. 
Both qualitative and quantitative comparisons are shown in Figure~\ref{fig:Comparisons} and Table~\ref{tab:quantitative_comparison}.
Our method, while introducing region-level retouching, maintains or even surpasses the global retouching performance of existing state-of-the-art methods in terms of objective evaluation, demonstrating the effectiveness of PerTouch.

\begin{figure}[t]
    \centering
    \vspace{-20pt}
    \includegraphics[width=\linewidth]{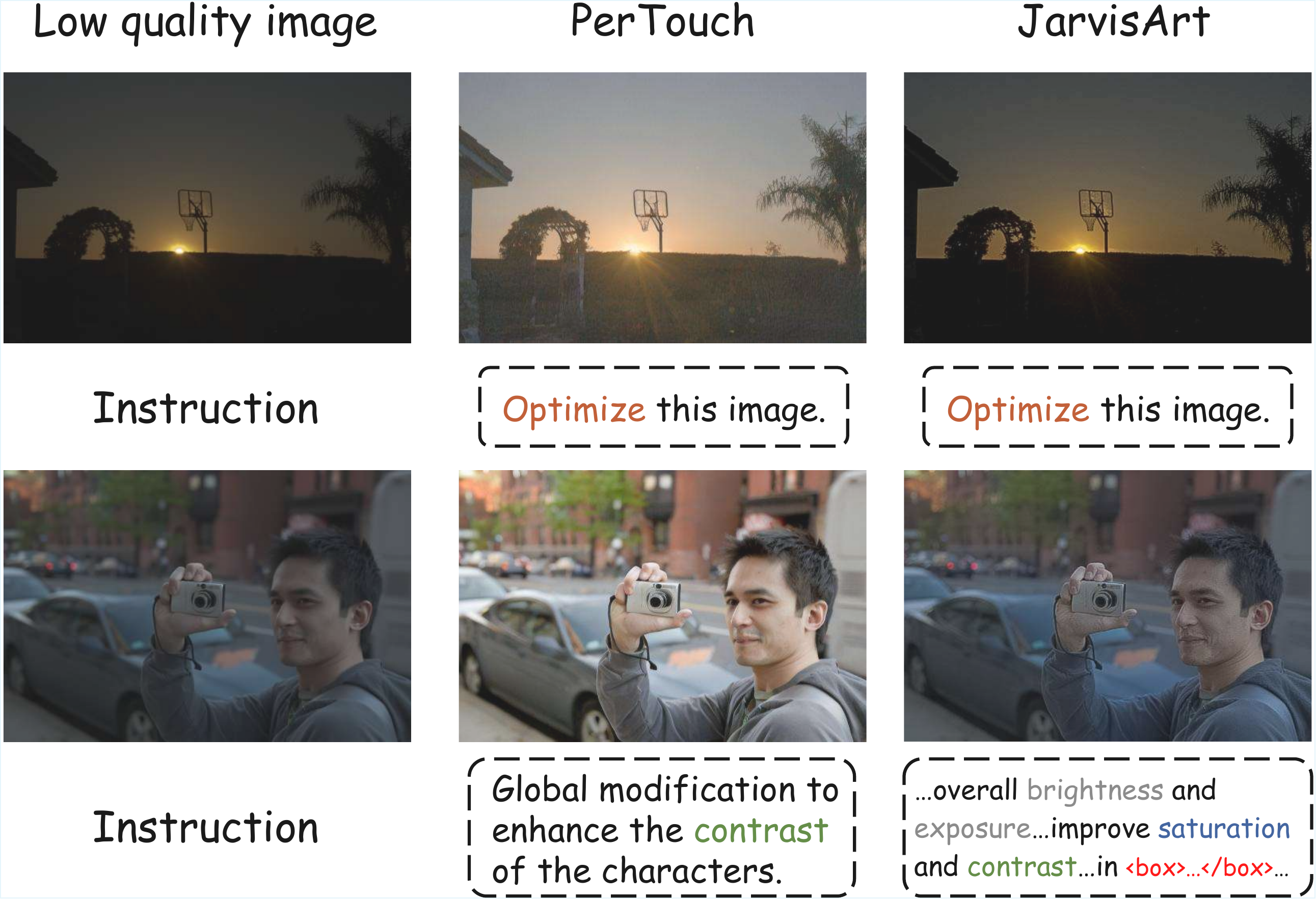}
    \setlength{\abovecaptionskip}{-3pt}
    \caption{Comparison with Jarvis Art.}
    \label{fig:Comparison w/ Jarvis}
\end{figure}

In addition, recent works explore using vision language models (VLMs) as agents to control photo-editing toolchains like Adobe Lightroom. 
However, the MIT-Adobe FiveK dataset does not contain detailed retouching path descriptions between low-quality inputs and ground-truth images, making direct quantitative evaluation of such systems infeasible. 
Therefore, we include a qualitative comparison for reference, as illustrated in Figure~\ref{fig:Comparison w/ Jarvis}.

\begin{figure}[t]
    \centering
    \vspace{-14pt}
    \includegraphics[width=0.9\linewidth]{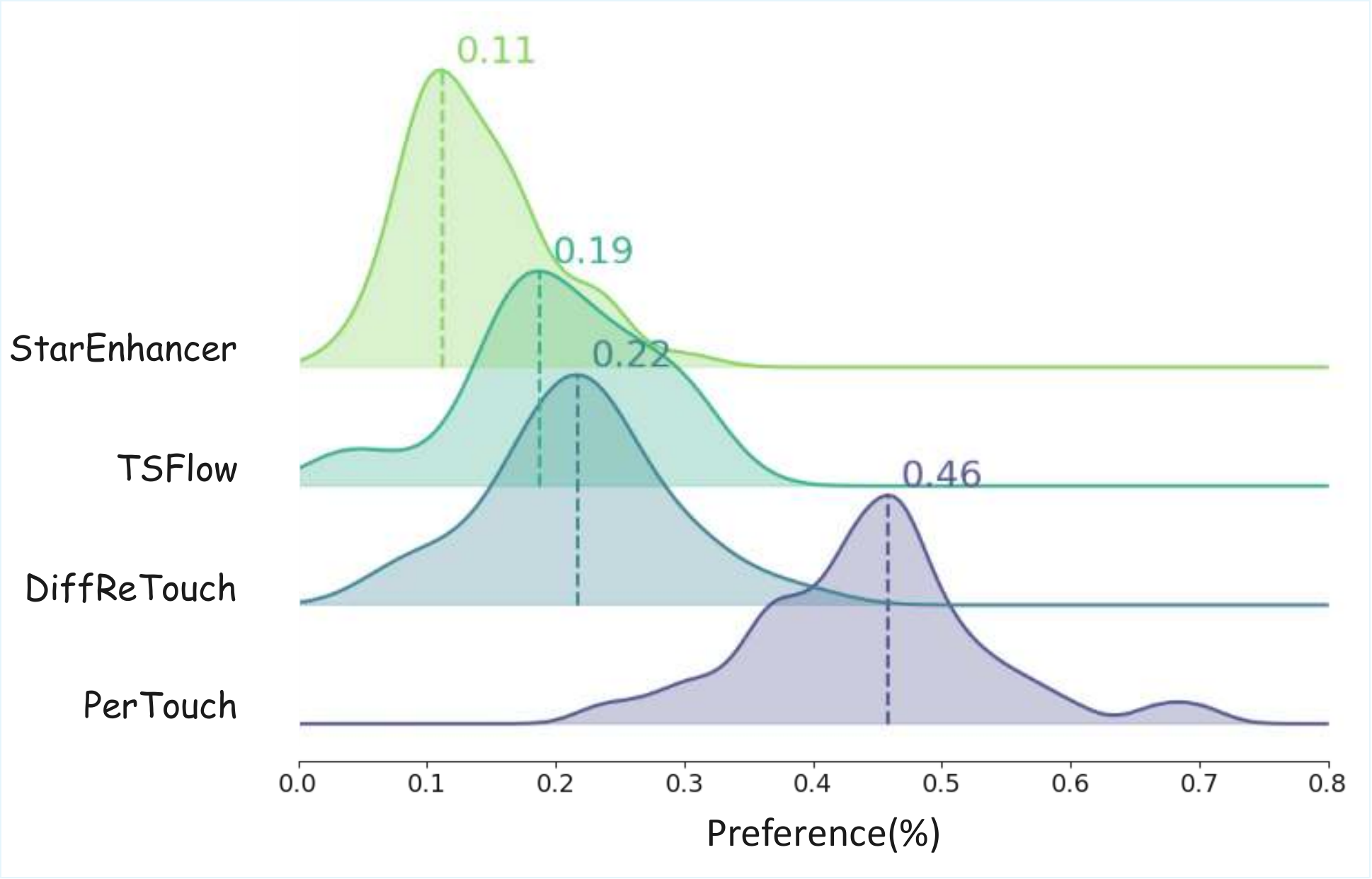}
    \setlength{\abovecaptionskip}{0pt}
    \caption{KDE plot of user study. Comparison of high-quality image selection rates.}
    \label{fig:User study}
    \vspace{-6pt}
\end{figure}

To further validate the effectiveness of our approach, we conduct a user study to assess human preferences over PerTouch and other SOTA baselines including DiffRetouch, StarEnhancer, TSFlow, and JarvisArt.
%
% Detailed results are presented and analyzed in the Supplementary Material.
We randomly select 30 images from the MIT-Adobe FiveK test set and recruit 50 volunteers to participate in the evaluation. 
Given the original input and retouched results from all methods, participants are asked to choose the result that best aligns with their personal preferences. 
We calculated the preference percentage for each method per user and summarized the results in the Kernel Density Estimation plot shown in Figure~\ref{fig:User study}.
The majority of participants expressed a preference for our method in nearly half of the test cases, significantly outperforming other approaches. 
This demonstrates the strong ability of our method to generate visually pleasing and user-preferred results.

\begin{figure}[ht]
    \centering
    \vspace{-20pt}
    \includegraphics[width=0.9\linewidth]{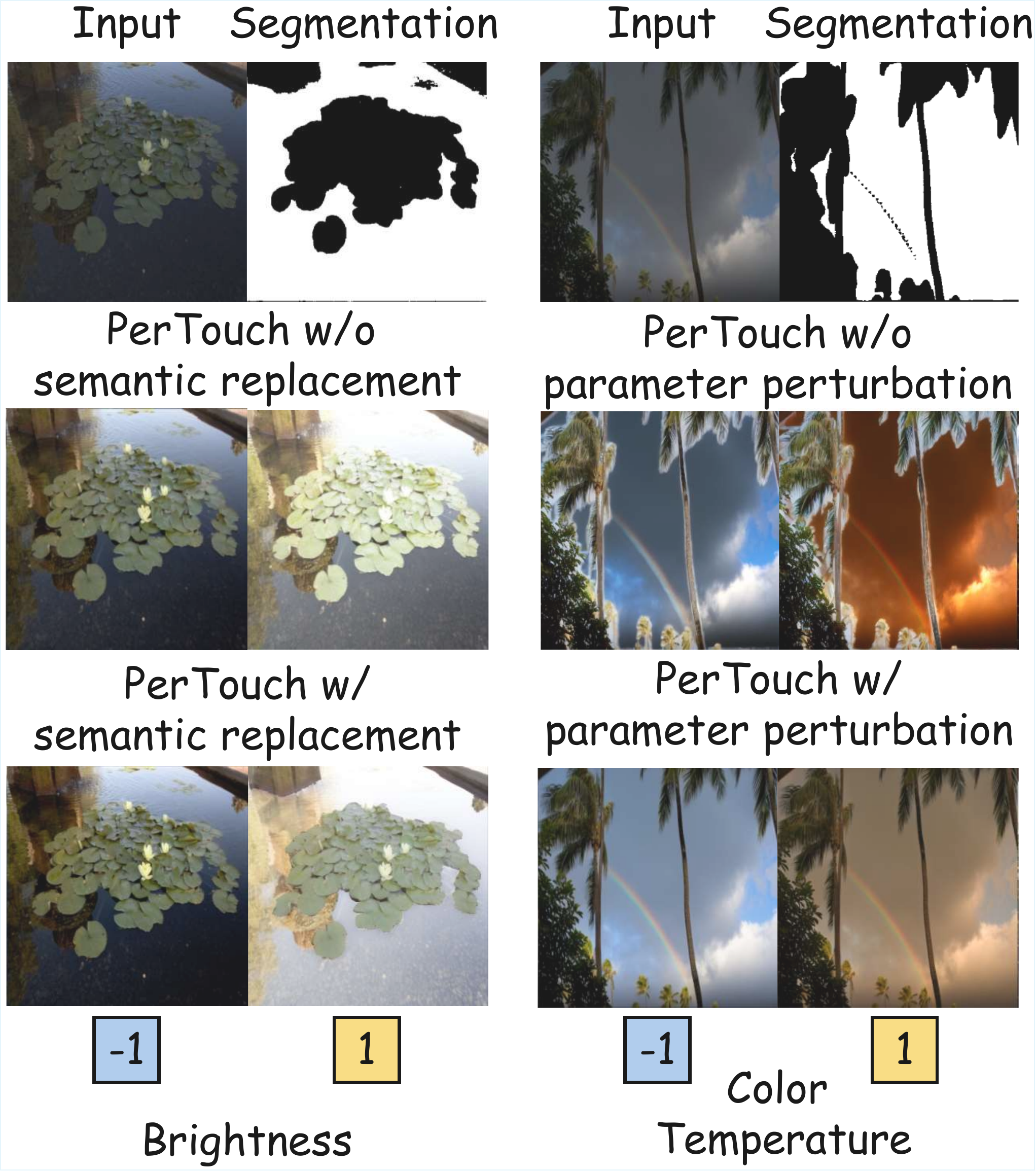}
    \setlength{\abovecaptionskip}{6pt}
    \caption{Ablation studies on key components of PerTouch. The left column compares results with and without the Semantic Replacement Module, showing its effectiveness in improving semantic region control and reducing undesired global spillover. The right column compares results with and without the Perturbation Mechanism, demonstrating its role in mitigating overfitting to segmentation boundaries and enhancing global visual quality. We only set the parameter values of the masked region dimensions to 1/-1, leaving all others to their default values.}
    \label{fig:Ablation study}
    \vspace{-7pt}
\end{figure}

\subsection{Ablation Studies}
\label{sec:ablation}
\subsubsection{Semantic Replacement Module} 

This module is designed to enhance the model's understanding of semantic regions within the image and improve the accuracy and consistency of semantic-aware local retouching. 
We observe that the provided parameter maps are often spatially discrete, while real images exhibit natural spatial continuity. 
Directly injecting such discrete control signals into the model often leads to ambiguity around semantic boundaries. 
%
% To address this issue, we introduce a semantic replacement module, which constructs training samples with consistent semantics but different spatial structures, encouraging the model to learn more robust regional awareness. 
%
In the ablation experiment, we removed this module while keeping all other settings unchanged and retrained the model. 
As shown in the left column of Figure~\ref{fig:Ablation study}, the absence of the semantic replacement module significantly degrades the model's ability to localize retouching, leading to spillover effects, where local edits undesirably affect global regions. 
This confirms the necessity of the semantic replacement module in improving region-level control precision and generalization.

\subsubsection{Perturbation Mechanism}

We found that directly injecting parameter maps obtained after semantic replacement can cause the model to overly rely on externally encoded segmentation boundaries, resulting in overfitting to these semantic borders. 
This behavior deviates from our goal, which is to allow the model to internally balance the global aesthetic guided by the diffusion prior with the localized control suggested by segmentation cues. 
To this end, we introduce a perturbation mechanism that encourages the model to perceive how different parameter values influence semantic boundaries. 
As shown in the right column of Figure~\ref{fig:Ablation study}, removing this mechanism causes the model to overfit external segmentation structures, resulting in reduced global visual coherence during user-guided manipulation. 
This demonstrates the importance of the perturbation mechanism in enhancing semantic awareness and improving user experience.

\section{Conclusion}

In this paper, we propose PerTouch, a unified diffusion-based framework for personalized image retouching. 
By introducing an explicit image-to-parameter mapping mechanism, along with semantic replacement and parameter perturbation modules, our method enables fine-grained, region-aware image retouching. 
To further align with user intent, we incorporate an agent that supports prompt-based control, iterative feedback refinement, and long-term preference modeling through scene-aware memory. 
Extensive experiments validate the effectiveness of each component and demonstrate that PerTouch generates high-quality results consistent with user preferences.

\section*{Acknowledgments}

\bigskip
\noindent This work was supported in part by the National Natural Science Foundation of China (62306153, 62225604), the Natural Science Foundation of Tianjin, China (24JCJQJC00020), the Young Elite Scientists Sponsorship Program by CAST (YESS20240686), the Fundamental Research Funds for the Central Universities (Nankai University, 070-63243143),  and Shenzhen Science and Technology Program (JCYJ20240813114237048). 
This work was also funded by Samsung R\&D Institute China-Beijing (SRC-B).
The computational devices is supported by the Supercomputing Center of Nankai University (NKSC).

% Note: \bibliographystyle{aaai2026} is automatically set by aaai2026.sty
% Do not add \bibliographystyle{aaai2026} here as it will cause "Illegal, another \bibstyle command" error
\bibliography{aaai2026}

@String(PAMI = {IEEE Trans. Pattern Anal. Mach. Intell.})

@String(CVPR= {IEEE Conf. Comput. Vis. Pattern Recog.})

@String(ICCV= {Int. Conf. Comput. Vis.})

@String(ECCV= {Eur. Conf. Comput. Vis.})

@String(ICPR = {Int. Conf. Pattern Recog.})

@String(TOG= {ACM Trans. Graph.})

@String(TIP  = {IEEE Trans. Image Process.})

@String(ICME = {Int. Conf. Multimedia and Expo})

@String(AAAI = {AAAI})

@String(PAMI  = {IEEE TPAMI})

@String(CVPR  = {CVPR})

@String(ICCV  = {ICCV})

@String(ECCV  = {ECCV})

@String(ICPR  = {ICPR})

@String(TOG   = {ACM TOG})

@String(TIP   = {IEEE TIP})

@String(ICME  =	{ICME})

@INPROCEEDINGS{5995332,
  author={Bychkovsky, Vladimir and Paris, Sylvain and Chan, Eric and Durand, Fredo},
  booktitle=CVPR, 
  title={Learning photographic global tonal adjustment with a database of input / output image pairs}, 
  year={2011},
  pages={97-104},
}

@inproceedings{jie2021PPR10K,
  title={PPR10K: A Large-Scale Portrait Photo Retouching Dataset with Human-Region Mask and Group-Level Consistency},
  author={Liang, Jie and Zeng, Hui and Cui, Miaomiao and Xie, Xuansong and Zhang, Lei},
  booktitle=CVPR,
  year={2021},
  pages={653--661},
}

@article{Chen2018LearningTS,
  title={Learning to See in the Dark},
  author={Cheng Chen and Qifeng Chen and Jia Xu and Vladlen Koltun},
  journal=CVPR,
  year={2018},
  pages={3291--3300},
}

@INPROCEEDINGS{8578758,
  author={Chen, Yu-Sheng and Wang, Yu-Ching and Kao, Man-Hsin and Chuang, Yung-Yu},
  booktitle=CVPR, 
  title={Deep Photo Enhancer: Unpaired Learning for Image Enhancement from Photographs with GANs}, 
  year={2018},
  pages={6306--6314},
}

@article{He2020ConditionalSM,
  title={Conditional Sequential Modulation for Efficient Global Image Retouching},
  author={Jingwen He and Yihao Liu and Y. Qiao and Chao Dong},
  journal={ArXiv},
  year={2020},
  pages={679--695},
}

@INPROCEEDINGS{9710400,
  author={Kim, Hanul and Choi, Su-Min and Kim, Chang-Su and Koh, Yeong Jun},
  booktitle=ICCV, 
  title={Representative Color Transform for Image Enhancement}, 
  year={2021},
  pages={4439-4448},
}

@article{Sun2021EnhanceIA,
  title={Enhance Images as You Like with Unpaired Learning},
  author={Xiaopeng Sun and Muxingzi Li and Tianyu He and Lubin Fan},
  journal={ArXiv},
  year={2021},
}

@INPROCEEDINGS{9577287,
  author={Liu, Risheng and Ma, Long and Zhang, Jiaao and Fan, Xin and Luo, Zhongxuan},
  booktitle=CVPR, 
  title={Retinex-inspired Unrolling with Cooperative Prior Architecture Search for Low-light Image Enhancement}, 
  year={2021},
  pages={10561--10570},
}

@InProceedings{Wang_2019_CVPR,
    author = {Wang, Ruixing and Zhang, Qing and Fu, Chi-Wing and Shen, Xiaoyong and Zheng, Wei-Shi and Jia, Jiaya},
    title = {Underexposed Photo Enhancement Using Deep Illumination Estimation},
    booktitle = CVPR,
    year = {2019},
    pages={6842-6850},
}

@INPROCEEDINGS{9102962,
  author={Zhu, Anqi and Zhang, Lin and Shen, Ying and Ma, Yong and Zhao, Shengjie and Zhou, Yicong},
  booktitle=ICME, 
  title={Zero-Shot Restoration of Underexposed Images via Robust Retinex Decomposition}, 
  year={2020},
  pages={1-6},
}

@article{Wang2021RealtimeIE,
  title={Real-time Image Enhancer via Learnable Spatial-aware 3D Lookup Tables},
  author={Tao Wang and Yong Li and Jingyang Peng and Yipeng Ma and Xian Wang and Fenglong Song and Youliang Yan},
  journal=ICCV,
  year={2021},
  pages={2471--2480},
}

@article{Yang2022AdaIntLA,
  title={AdaInt: Learning Adaptive Intervals for 3D Lookup Tables on Real-time Image Enhancement},
  author={Canqian Yang and Meiguang Jin and Xu Jia and Yi Xu and Ying Chen},
  journal=CVPR,
  year={2022},
  pages={17522--17531},
}

@article{Zeng2020LearningI3,
  title={Learning Image-Adaptive 3D Lookup Tables for High Performance Photo Enhancement in Real-Time},
  author={Huiyu Zeng and Jianrui Cai and Lida Li and Zisheng Cao and Lei Zhang},
  journal=PAMI,
  year={2020},
  pages={2058--2073},
}

@INPROCEEDINGS{moran2020curl,
  author={Moran, Sean and McDonagh, Steven and Slabaugh, Gregory},
  booktitle=ICPR, 
  title={CURL: Neural Curve Layers for Global Image Enhancement}, 
  year={2021},
  pages={9796--9803},
}

@inproceedings{song2021starenhancer,
  title={StarEnhancer: Learning Real-Time and Style-Aware Image Enhancement},
  author={Song, Yuda and Qian, Hui and Du, Xin},
  booktitle=ICCV,
  year={2021},
  pages={4126--4135},
}

@article{Gharbi2017DeepBL,
  title={Deep bilateral learning for real-time image enhancement},
  author={Micha{\"e}l Gharbi and Jiawen Chen and Jonathan T. Barron and Samuel W. Hasinoff and Fr{\'e}do Durand},
  journal=TOG,
  year={2017},
  pages={1--12},
}

@InProceedings{Moran_2020_CVPR,
    author = {Moran, Sean and Marza, Pierre and McDonagh, Steven and Parisot, Sarah and Slabaugh, Gregory},
    title = {DeepLPF: Deep Local Parametric Filters for Image Enhancement},
    booktitle = CVPR,
    year = {2020},
    pages={12823-12832},
}

@inproceedings{kim2020pienet,
  title = {PieNet: Personalized Image Enhancement},
  author = {Kim, Han-Ul and Koh, Young Jun and Kim, Chang-Su},
  booktitle=ECCV,
  year = {2020},
  pages={374-390},
}

@ARTICLE{10225702,
  author={Kim, Heewon and Lee, Kyoung Mu},
  journal=TIP, 
  title={Learning Controllable ISP for Image Enhancement}, 
  year={2024},
  pages={867-880},
}

@inproceedings{duan2025diffretouch,
  title={DiffRetouch: Using Diffusion to Retouch on the Shoulder of Experts},
  author={Duan, Zheng-Peng and Zhang, Jiawei and Lin, Zheng and Jin, Xin and Wang, XunDong and Zou, Dongqing and Guo, Chun-Le and Li, Chongyi},
  booktitle=AAAI,
  year={2025},
  pages={2825--2833},
}

@article{oywq2023rsfnet,
  title={RSFNet: A white-Box image retouching approach using region-specific color filters},
  author={Wenqi Ouyang and Yi Dong and Xiaoyang Kang and Peiran Ren and Xin Xu and Xuansong Xie},
  journal={https://arxiv.org/abs/2303.08682},
  year={2023},
  pages={12160--12169},
}

@article{li2025hybridagents,
  title={Hybrid Agents for Image Restoration},
  author={Li, Bingchen and Li, Xin and Lu, Yiting and Chen, Zhibo},
  journal={arXiv preprint arXiv:2503.10120},
  year={2025},
}

@misc{chen2024restoreagent,
  title={RestoreAgent: Autonomous Image Restoration Agent via Multimodal Large Language Models},
  author={Chen, Haoyu and Li, Wenbo and Gu, Jinjin and Ren, Jingjing and Chen, Sixiang and Ye, Tian and Pei, Renjing and Zhou, Kaiwen and Song, Fenglong and Zhu, Lei},
  year={2024},
  eprint={2407.18035},
  archivePrefix={arXiv},
}

@misc{zhu2024agenticir,
  title={An Intelligent Agentic System for Complex Image Restoration Problems},
  author={Zhu, Kaiwen and Gu, Jinjin and You, Zhiyuan and Qiao, Yu and Dong, Chao},
  year={2024},
  eprint={2410.17809},
  archivePrefix={arXiv},
}

@article{Jiang2025MultiAgentIR,
  title={Multi-Agent Image Restoration},
  author={Xuexing Jiang and Gehui Li and Bin Chen and Jian Zhang},
  journal={ArXiv},
  year={2025},
}

@article{chen2025photoartagent,
  title={PhotoArtAgent: Intelligent Photo Retouching with Language Model-Based Artist Agents},
  author={Chen, Haoyu and Tao, Keda and Wang, Yizao and Wang, Xinlei and Zhu, Lei and Gu, Jinjin},
  journal={arXiv preprint arXiv:2505.23130},
  year={2025}
}

@article{jarvisart2025,
    title={JarvisArt: Liberating Human Artistic Creativity via an Intelligent Photo Retouching Agent}, 
    author={Yunlong Lin and Zixu Lin and Kunjie Lin and Jinbin Bai and Panwang Pan and Chenxin Li and Haoyu Chen and Zhongdao Wang and Xinghao Ding and Wenbo Li and Shuicheng Yan},
    year={2025},
    journal={arXiv preprint arXiv:2506.17612}
}

@article{dutt2025monetgpt,
  title={MonetGPT: Solving Puzzles Enhances MLLMs’ Image Retouching Skills},
  author={Dutt, Niladri Shekhar and Ceylan, Duygu and Mitra, Niloy J.},
  journal={arXiv preprint arXiv:2505.06176},
  year={2025}
}

@article{Rombach2021HighResolutionIS,
  title={High-Resolution Image Synthesis with Latent Diffusion Models},
  author={Robin Rombach and A. Blattmann and Dominik Lorenz and Patrick Esser and Bj{\"o}rn Ommer},
  journal=CVPR,
  year={2021},
  pages={10684--10695},
}

@misc{zhang2023adding,
  title={Adding Conditional Control to Text-to-Image Diffusion Models}, 
  author={Lvmin Zhang and Anyi Rao and Maneesh Agrawala},
  booktitle=ICCV,
  year={2023},
  pages={3836--3847},
}

@misc{AdobePhotoshop,
  author = {{Adobe Inc.}},
  title = {{Adobe Photoshop}},
  year = {2024},
  note = {Available at https://www.adobe.com/products/photoshop.html}
}

@misc{AdobeLightroom,
  author = {{Adobe Inc.}},
  title = {{Adobe Lightroom}},
  year = {2024},
  note = {Available at https://www.adobe.com/products/photoshop-lightroom.html}
}

\clearpage


\section*{Supplementary material}

# PerTouch: VLM-Driven Agent for Personalized and Semantic Image Retouching

—————— **Supplementary Material** ——————

Our supplementary material provides additional details about our method and experimental results, summarized as follows:

- Detailed structure of our PerTouch in Section 1.
- Detailed experiment setting in our training process in Section 2.
- Various supplementary experiments compared with other methods in Section 3.
- Limitation of PerTouch in Section 4.

## Detailed Structure

### Detailed Baseline Architecture

Our method is built upon Stable Diffusion, which operates in a learned latent space rather than directly in pixel space, extending the denoising diffusion probabilistic models (DDPM). A powerful autoencoder, consisting of an encoder $E$ and decoder $D$, is first pre-trained to map input images $X$ into latent representations $Z = E(X)$, and reconstruct them via $D(Z)$.

The denoising network $\varepsilon_\theta(Z_t, t, m)$ is trained to reverse the noise process in the latent space, where $Z_t$ denotes the noisy latent at timestep $t$, and $m$ denotes conditioning signals. In our use case, we do not rely on additional semantic prompts and instead apply empty textual embeddings as conditions, thereby emphasizing the inherent generative priors of the diffusion model.

To enable localized control over the generation process, we incorporate ControlNet into the Stable Diffusion pipeline. ControlNet augments the original U-Net backbone with a parallel control branch $\mathcal{F}_{\text{ctrl}}$, which receives spatial structural conditions $C \in \mathbb{R}^{H \times W \times d}$. This branch is composed of zero-initialized convolutional residual blocks and generates control features $\Delta h_t = \mathcal{F}_{\text{ctrl}}(C, t)$ at each timestep $t$. These control features are injected into the main U-Net backbone via residual addition:

$$h'_t = h_t + \Delta h_t. \tag{1}$$

During training, the parameters of the main denoising network $\theta$ remain frozen to preserve the learned generative priors, and only the parameters $\phi$ of the control branch are optimized.

### Detailed parameter map injection structure

To support multi-attribute regional control, we extend region-level attribute scores into a multi-channel spatial conditioning map $C = \{C^1, C^2, \ldots, C^K\}$, where each channel $C^k \in \mathbb{R}^{H \times W}$ represents the spatial distribution of a specific image attribute (e.g., colorfulness, contrast, color temperature, brightness). These maps are aligned with the resolution of the input segmentation mask. Each pixel in channel $C^k$ reflects the attribute score assigned to the corresponding region in the original image. These spatial attribute maps serve as fine-grained control signals and are fed into ControlNet's control branch at each timestep. This design allows the model to learn region-aware generation behavior while preserving the global image quality ensured by Stable Diffusion. During inference, users can manipulate the visual style of different regions by adjusting the values in the respective attribute maps. The coefficient values in each channel are normalized to the range $[-1, 1]$, representing the learned distribution of high-quality image styles. For example, in the brightness channel, higher values induce brighter appearances, while lower values yield darker tones. This formulation enables personalized, attribute-specific image retouching in a spatially controlled manner. Training is conducted using frozen weights in the main U-Net and optimizing only the ControlNet branch. Detailed training and inference architectures are illustrated in Figure 1 and Figure 2

### Detailed Construction of Image-to-Parameter Pairs

To support fine-grained control over image attributes, we construct a four-channel parameter map $C = \{C^1, C^2, C^3, C^4\} \in \mathbb{R}^{4 \times H \times W}$, where each channel $C^k \in \mathbb{R}^{H \times W}$ encodes the spatial distribution of a specific perceptual attribute: colorfulness, contrast, color temperature, and brightness, respectively.

Each attribute map $C^k$ is formed by computing a region-wise scalar score $C^k(R_i)$ for each segmented region $R_i$, and broadcasting this score to all pixels within the region. Region masks are generated using SAM2, and only sufficiently large and semantically meaningful segments are retained for scoring. The computation methods for $C^k(R_i)$ are defined as follows:

**Colorfulness.** Let $rg = |R - G|$, $yb = |0.5(R + G) - B|$ denote opponent color channels. The region-level colorful-

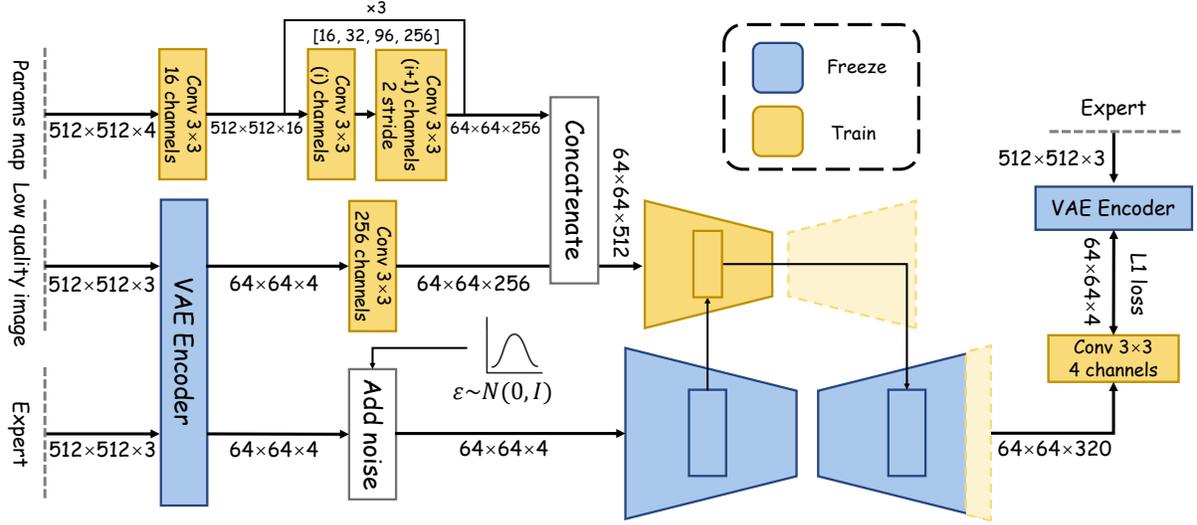

Figure 1: Detailed Training Architecture

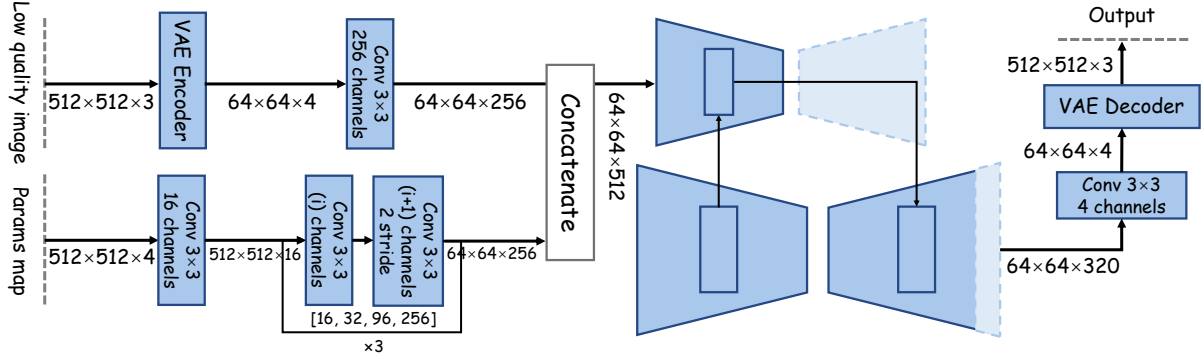

Figure 2: Detailed Inference Architecture

ness is:

$$C^1(R_i) = \sqrt{\sigma_{rg}^2 + \sigma_{yb}^2} + 0.3 \cdot \sqrt{\mu_{rg}^2 + \mu_{yb}^2}, \quad (2)$$

where $\mu$ and $\sigma$ are computed over pixels in $R_i$.

**Contrast.** Defined as the average local color difference in the region:

$$C^2(R_i) = \frac{1}{4N} \sum_{x_i \in R_i} \sum_{x_j \in \mathcal{N}(x_i)} \|x_i - x_j\|_2^2, \quad (3)$$

where $\mathcal{N}(x_i)$ are the 4-neighbors of pixel $x_i$ within the region.

**Color Temperature.** After converting the mean RGB color in $R_i$ into CIE chromaticity coordinates $(x, y)$, we estimate CCT using the Hernandez-Andres model. To reflect perceptual warmth, we take the negative value:

$$C^3(R_i) = -\text{CCT}(x, y), \quad (4)$$

to reflect perceptual warmth.

**Brightness.** Calculated using a standard perceptual brightness formula:

$$C^4(R_i) = \sqrt{0.241 \cdot \bar{R}^2 + 0.691 \cdot \bar{G}^2 + 0.068 \cdot \bar{B}^2}, \quad (5)$$

where $\bar{R}, \bar{G}, \bar{B}$ are average color values in regions $R_i$.

The four maps $\{C^1, \ldots, C^4\}$ are constructed by assigning each scalar $C^k(R_i)$ to all pixels in $R_i$, and then normalized to $[-1, 1]$ based on global dataset statistics. This provides a spatially aligned, attribute-specific guidance signal for regional image generation.

### Detailed Semantic Replacement Module

To mitigate over-reliance on parameter maps and encourage the model to perceive semantic boundaries, we introduce a region-level semantic replacement strategy. Given a source image and an expert-retouched target, we first compute the attribute vector $C(R_i) = [C^1(R_i), C^2(R_i), C^3(R_i), C^4(R_i)]$ for each segmented region $R_i$. We then identify the region $R^*$ with the greatest attribute discrepancy:

$$R^* = \arg\max_{R_i} \|C_{\text{src}}(R_i) - C_{\text{tgt}}(R_i)\|_2. \quad (6)$$

We replace both the pixels and corresponding parameter map of region $R^*$ in the source image with those from the target image. This synthetic perturbation introduces local inconsistencies that encourage the model to detect region-aware changes, thereby improving its ability to learn fine-grained, spatially controllable retouching behaviors.

## Experiment Setting

### Datasets

Our experiments are conducted on the MIT-Adobe FiveK dataset (Bychkovsky et al. 2011), which contains 5,000 RAW images, each accompanied by five expert-retouched versions (A/B/C/D/E). We follow the preprocessing pipeline of MIT-Adobe-5K-UPE and split the dataset into 4,500 pairs for training and 500 pairs for validation (Song, Qian, and Du 2021; Wang et al. 2019). To adapt the dataset to our model's requirements, we construct image–parameter map pairs for each sample using the data preprocessing pipeline proposed in Section 3.2 in our main text.

### Train Details

Our proposed PreTouch is built upon Stable Diffusion 2.1-base (Rombach et al. 2021), added with ControlNet (Zhang, Rao, and Agrawala 2023) for conditional guidance. We initialize our model using the pretrained Stable Diffusion weights, and follow the standard ControlNet approach by duplicating the encoder weights from Stable Diffusion while inserting zero convolutions in the injection layers to prevent interference with the backbone network. Most of the parameters of Stable Diffusion are frozen during training; only the ControlNet modules and the final layer of the UNet are updated. This design helps preserve the generative prior of the diffusion model while improving the model's ability to reconstruct fine image details. We train the model for approximately 40 epochs with a batch size of 7. The AdamW (Loshchilov and Hutter 2017) optimizer is used with a learning rate of $5 \times 10^{-5}$. All experiments are conducted on six NVIDIA GeForce RTX 3090 24GB GPUs. During training, input images are resized to $512 \times 512$ before being fed into the network. After encoding by the VAE provided by Stable Diffusion, the latent representations used in the diffusion process have a spatial resolution of $64 \times 64$ with 4 channels. During inference, we adopt the improved DDPM sampling strategy with 50 timesteps, consistent with the original Stable Diffusion setup. The input image $R$ is encoded into the latent space via the same VAE and concatenated with the noise latent $Z$. The replacement rate $\alpha$ in the semantic replacement module is set to 0.05 for the first about 35 epochs and 0 for the rest. In the perturbation module of the parameter map, the maximum random shift $S$ is set to 20.0, and the maximum Gaussian blur $\sigma$ is set to 5.0. The loss balancing hyperparameters $\lambda$ and $\beta$ are set to 1 and 0.01, respectively.

For the Agent component, in order to balance generation speed and semantic understanding, our PerTouch adopts Qwen-2.5-VL for instruction following, instruction parsing, and region grounding tasks, while utilizing QVQ-Max with chain-of-thought (CoT) capabilities in the Feedback-driven

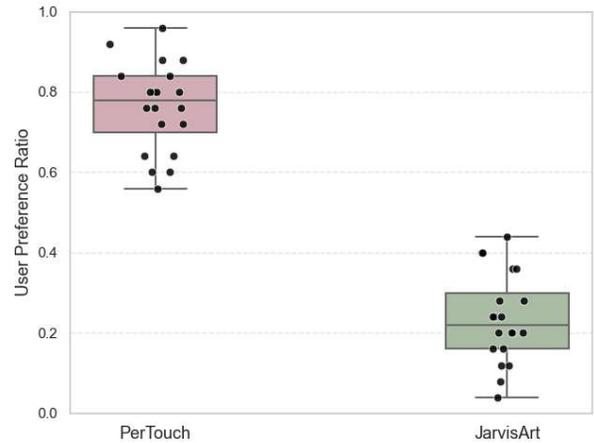

Figure 3: User study with JarvisArt.

Rethinking mechanism. Since our method is entirely train-free, it can be conveniently deployed through API calls without additional model tuning or fine-tuning procedures.

## More Experiments

### More Qualitative Results

To complement the qualitative comparisons presented in the main text, we provide additional visual results in Figure 6 and Figure 7. These examples further demonstrate PerTouch's ability to generate expert-specific retouching outputs that align with the diverse preferences exhibited in the MIT-Adobe FiveK dataset. For each input image, we present retouching outputs corresponding to all five expert styles. Compared to other methods, PerTouch achieves better color consistency and faithfully captures and restores the subtle differences in each expert's style.

### More Comparisons with JarvisArt

To further supplement the qualitative analysis with JarvisArt presented in the main paper, we provide additional visual comparisons in this section, showcasing the strengths of our method in both global and local retouching scenarios. Due to the absence of step-wise retouching annotations in the MIT-Adobe FiveK dataset, these results are intended solely for visual reference and not for quantitative benchmarking.

For global retouching, we uniformly adopt the prompt "Globally retouch this image." across all comparisons. In the case of local retouching, we use our Strong Instruction for PerTouch, while JarvisArt is given more detailed and descriptive instructions to ensure the best possible performance. The exact prompts used are illustrated in Figure 5. The results demonstrate that our PerTouch consistently delivers aesthetically pleasing outcomes in both global and local retouching tasks. In contrast, JarvisArt often produces unstable outputs that lack visual appeal.

To further validate these observations, we conducted a user study. We randomly selected 20 images from the MIT-Adobe FiveK test set and used the same prompt "Globally retouch this image." to generate outputs from both PerTouch

Low quality image

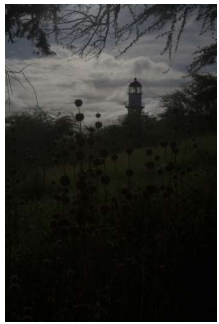

"Significantly increase brightness"

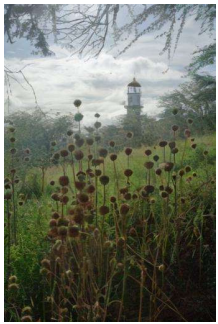

w/o Scene-aware Memory

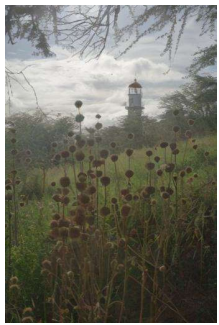

w/o Feedback-driven Rethinking

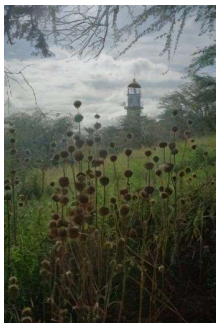

Figure 4: Ablation Study.

and JarvisArt. Fifty participants were then asked to select the result that better matched their aesthetic preferences. As shown in Figure 3, a clear majority of users favored the results produced by our method, confirming the visual superiority of PerTouch.

## More Ablation Study

To further validate the effectiveness of our Agent design, we conduct additional ablation studies on the Scene-Aware Memory and Feedback-Driven Rethinking mechanism, as shown in Figure 4.

We visualize the retouching results under three conditions: (1) without scene-aware memory, (2) without the rethinking mechanism, and (3) with both mechanisms enabled. As shown in the figure, for a weak instruction such as "Significantly increase the brightness," removing the scene-aware memory prevents the system from recalling the user's historical preference for high saturation in similar scenes (in this case, we added a preference for higher saturation under this scene in the user history). As a result, the retouched output appears bland and fails to adjust the saturation accordingly. On the other hand, disabling the rethinking mechanism results in insufficient editing of the brightness attribute, failing to fulfill the "significant" requirement implied by the user. These outcomes suggest that the model struggles to ro-

bustly interpret vague instructions without proper feedback refinement and long-term preference modeling.

In contrast, the full model equipped with both modules produces more consistent and aesthetically pleasing results that better align with user intent. These findings highlight the importance of both Scene-Aware Memory and Feedback-Driven Rethinking mechanisms in subjective image retouching.

## Limitation

The pretrained Stable Diffusion model possesses a strong generative prior, enabling it to produce globally appealing retouching results. However, due to information loss during the encoding-decoding process and the introduced randomness of the generative procedure, our outputs still exhibit texture artifacts in regions with intricate details, even after employing targeted training strategies. To alleviate this issue, inspired by existing methods (Duan et al. 2025; Gharbi et al. 2017), we attempted to incorporate an affine bilateral grid, which performs edge-aware transformations by applying spatially varying affine mappings across the image. Specifically, the image is divided into a fixed-size grid (e.g., $16 \times 16$), and each cell is assigned a learned affine transformation that is then interpolated over the entire image. While this structure can achieve efficient and edge-preserving enhancement, we found it insufficient for semantically meaningful, fine-grained retouching. This is primarily due to its reliance on low-resolution grids that do not align well with semantic boundaries, limiting its expressiveness and controllability. In future work, we aim to explore alternative architectural designs that better balance edge-preserving properties and local generative flexibility, allowing for more accurate and coherent retouching in complex textured regions.

## References

Bychkovsky, V.; Paris, S.; Chan, E.; and Durand, F. 2011. Learning photographic global tonal adjustment with a database of input / output image pairs. In *CVPR*, 97–104.

Duan, Z.-P.; Zhang, J.; Lin, Z.; Jin, X.; Wang, X.; Zou, D.; Guo, C.-L.; and Li, C. 2025. DiffRetouch: Using Diffusion to Retouch on the Shoulder of Experts. In *AAAI*, 2825–2833.

Gharbi, M.; Chen, J.; Barron, J. T.; Hasinoff, S. W.; and Durand, F. 2017. Deep bilateral learning for real-time image enhancement. *ACM TOG*, 1–12.

Loshchilov, I.; and Hutter, F. 2017. Fixing Weight Decay Regularization in Adam. *ArXiv*.

Rombach, R.; Blattmann, A.; Lorenz, D.; Esser, P.; and Ommer, B. 2021. High-Resolution Image Synthesis with Latent Diffusion Models. *CVPR*, 10684–10695.

Song, Y.; Qian, H.; and Du, X. 2021. StarEnhancer: Learning Real-Time and Style-Aware Image Enhancement. In *ICCV*, 4126–4135.

Wang, R.; Zhang, Q.; Fu, C.-W.; Shen, X.; Zheng, W.-S.; and Jia, J. 2019. Underexposed Photo Enhancement Using Deep Illumination Estimation. In *CVPR*, 6842–6850.

Zhang, L.; Rao, A.; and Agrawala, M. 2023. Adding Conditional Control to Text-to-Image Diffusion Models.

Low quality image       PerTouch       JarvisArt

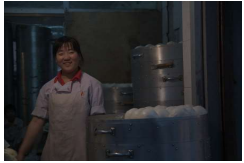 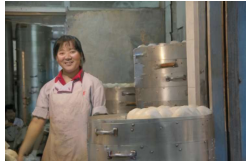 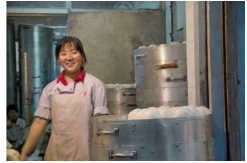 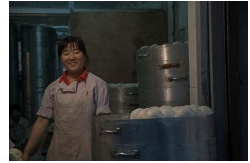 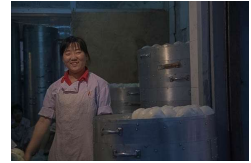

Global retouching this image.

Improve saturation of the character.

Global retouching this image.

Improve the overall exposure and saturation of the image to ensure the beauty of the image. In the region <box>0.038, 0.195, 0.471, 1.000</box> , Increase the saturation of the character area.

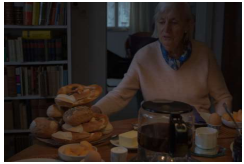 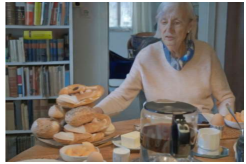 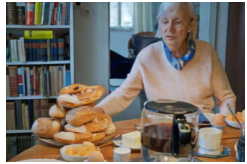 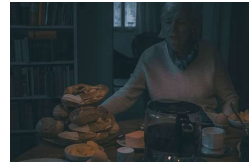 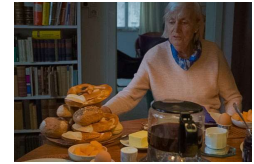

Global retouching this image.

Improve overall contrast.

Global retouching this image.

Improve global brightness, increase global saturation, and improve global contrast.

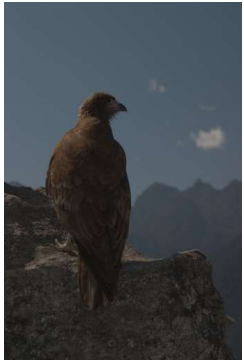 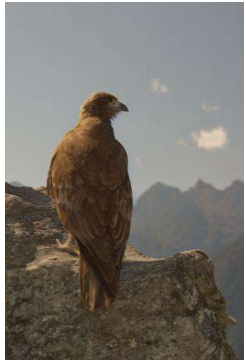 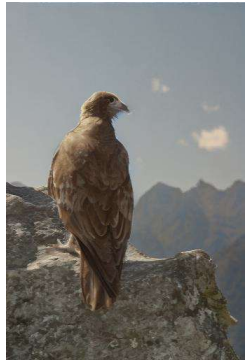 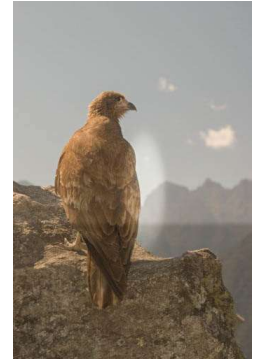 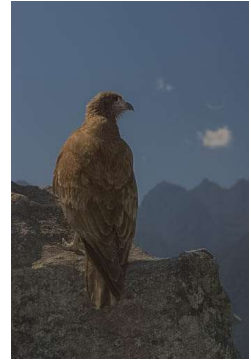

Global retouching this image.

Significantly increased eagle brightness.

Global retouching this image.

Improve the overall brightness, ensure the overall beauty of the image, and moderately increase the saturation and contrast. In the region <box>0.117, 0.225, 0.607, 0.914</box>, Greatly increases the brightness of the Eagle.

Figure 5: More comparisons with JarvisArt.

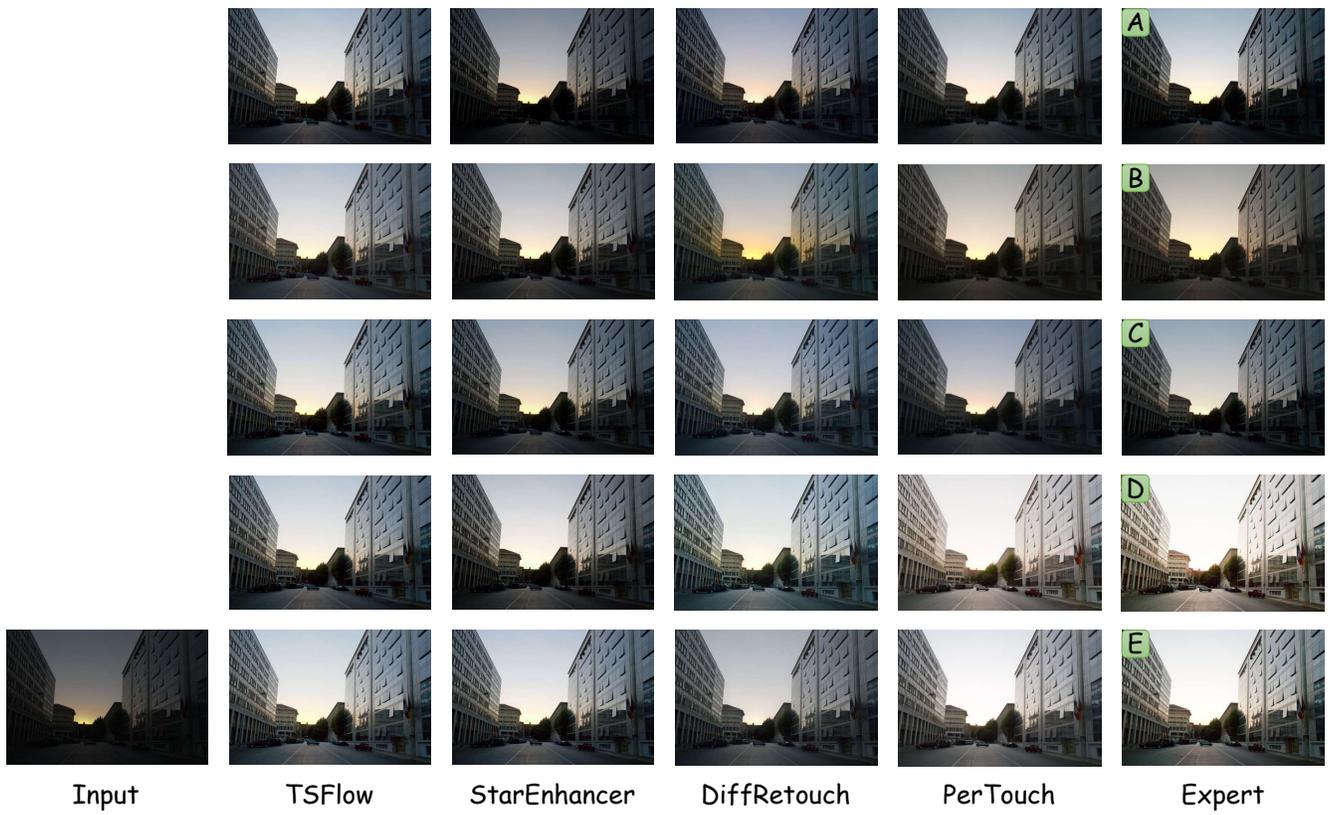

|        |        |              |             |          |        |
|--------|--------|--------------|-------------|----------|--------|
| Input  | TSFlow | StarEnhancer | DiffRetouch | PerTouch | Expert |

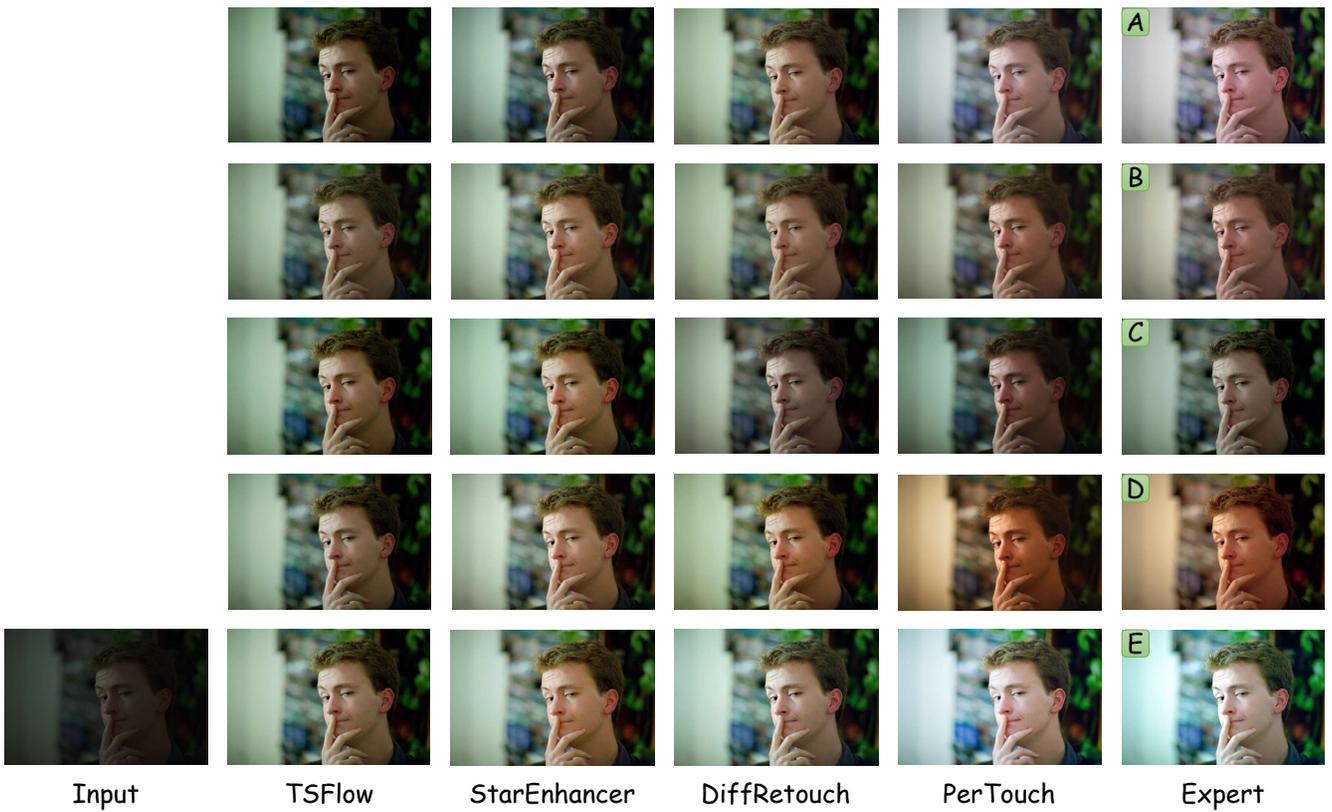

|        |        |              |             |          |        |
|--------|--------|--------------|-------------|----------|--------|
| Input  | TSFlow | StarEnhancer | DiffRetouch | PerTouch | Expert |

Figure 6: More comparisons with other methods.

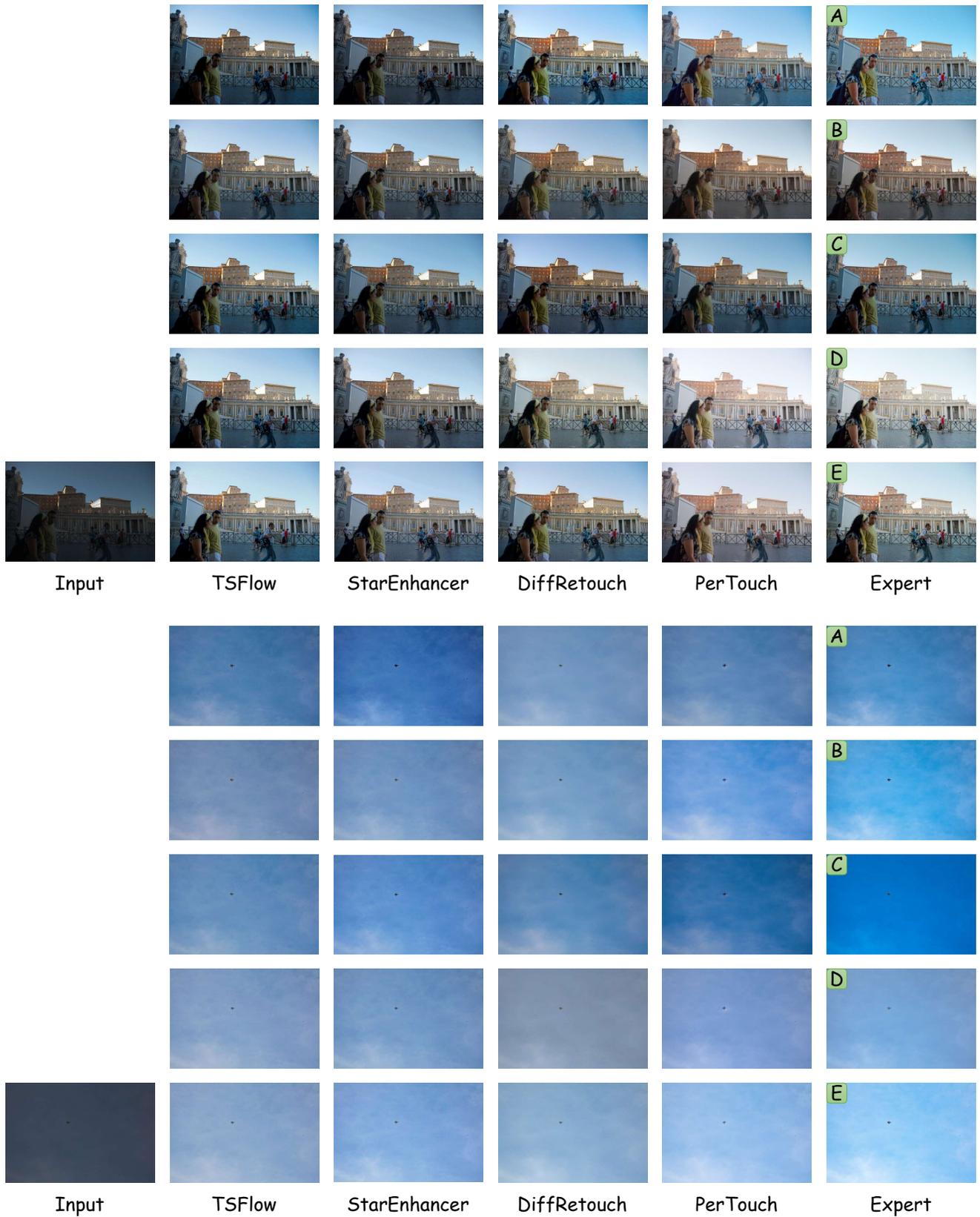

Figure 7: More comparisons with other methods.



\end{document}